%% file: icml2024.tex
\documentclass{article}
\usepackage[icml,subfig]{definition}
\usepackage[accepted]{icml2024}
\usepackage{titletoc}
\usepackage{mathrsfs}
\usepackage{upquote}
\usepackage[linewidth=0.5pt,leftmargin=1em,rightmargin=1em,innertopmargin=5pt,
    innerbottommargin=5pt,
    innerrightmargin=5pt,
    innerleftmargin=5pt]{mdframed}
\usepackage[toc,page,header]{appendix}

\setlist{leftmargin=*,nosep}

\newcommand{\ours}{\textsc{SciBench}\xspace}
\newcommand{\scite}[1]{{\scriptsize\cite{#1}}}

\begin{document}

\twocolumn[
\icmltitle{\ours: Evaluating College-Level Scientific Problem-Solving Abilities of Large Language Models}

\icmlsetsymbol{equal}{*}

\begin{icmlauthorlist}
\icmlauthor{Xiaoxuan Wang}{equal,ucla}
\icmlauthor{Ziniu Hu}{equal,caltech}
\icmlauthor{Pan Lu}{equal,ucla}
\icmlauthor{Yanqiao Zhu}{equal,ucla}
\icmlauthor{Jieyu Zhang}{uw}
\icmlauthor{Satyen Subramaniam}{ucla}
\icmlauthor{Arjun R. Loomba}{ucla}
\icmlauthor{Shichang Zhang}{ucla}
\icmlauthor{Yizhou Sun}{ucla}
\icmlauthor{Wei Wang}{ucla}
\end{icmlauthorlist}

\icmlaffiliation{ucla}{University of California, Los Angeles, Los Angeles, CA, USA}
\icmlaffiliation{uw}{University of Washington, Seattle, WA, USA}
\icmlaffiliation{caltech}{California Institute of Technology, Pasadena, CA, USA}

\icmlcorrespondingauthor{Xiaoxuan Wang}{xw27@cs.ucla.edu}

\icmlkeywords{Large Language Models, Scientific Problem Reasoning}
\vspace{-0.2em}
\begin{center}
Project Homepage: \url{https://scibench-ucla.github.io}
\end{center}
\vskip 0.15in
]

\printAffiliationsAndNotice{\icmlEqualContribution}
\begin{abstract}
Most existing Large Language Model (LLM) benchmarks on scientific problem reasoning focus on problems grounded in high-school subjects and are confined to elementary algebraic operations.
To systematically examine the reasoning capabilities required for solving complex scientific problems, we introduce an expansive benchmark suite \ours for LLMs.
\ours contains a carefully curated dataset featuring a range of collegiate-level scientific problems from mathematics, chemistry, and physics domains.
Based on the dataset, we conduct an in-depth benchmarking study of representative open-source and proprietary LLMs with various prompting strategies.
The results reveal that current LLMs fall short of delivering satisfactory performance, with the best overall score of merely 43.22\%. Furthermore, through a detailed user study, we categorize the errors made by LLMs into ten problem-solving abilities. Our analysis indicates that no single prompting strategy significantly outperforms the others and some strategies that demonstrate improvements in certain problem-solving skills could result in declines in other skills.
We envision that \ours will catalyze further developments in the reasoning abilities of LLMs, thereby ultimately contributing to scientific research and discovery.
\end{abstract}
\input{sections/intro.tex}

\input{sections/related-work.tex}

\input{sections/datasets.tex}
\input{sections/experiments.tex}
\input{sections/analysis.tex}

\input{sections/conclusion.tex}
\input{sections/reproducibility}
\input{sections/ethical}
\input{sections/impact}

\section*{Acknowledgements}
This work was supported by the National Science Foundation (NSF) under Grant Nos. 1829071, 1937599, 2106859, 2119643, 2202693, 2211557, 2303037, and 2312501; the National Institutes of Health (NIH) under Grant No. U54HG012517; the Defense Advanced Research Projects Agency (DARPA) under Grant No. HR00112490370; NASA; SRC JUMP 2.0 Center; Amazon Research Awards; and Snapchat Gifts.

\bibliographystyle{icml2024}
\bibliography{icml2024}

\input{sections/appendix}

\end{document}

%% file: sections/intro.tex
\section{Introduction}

\begin{figure*}
    \centering
    \includegraphics[width=\linewidth]{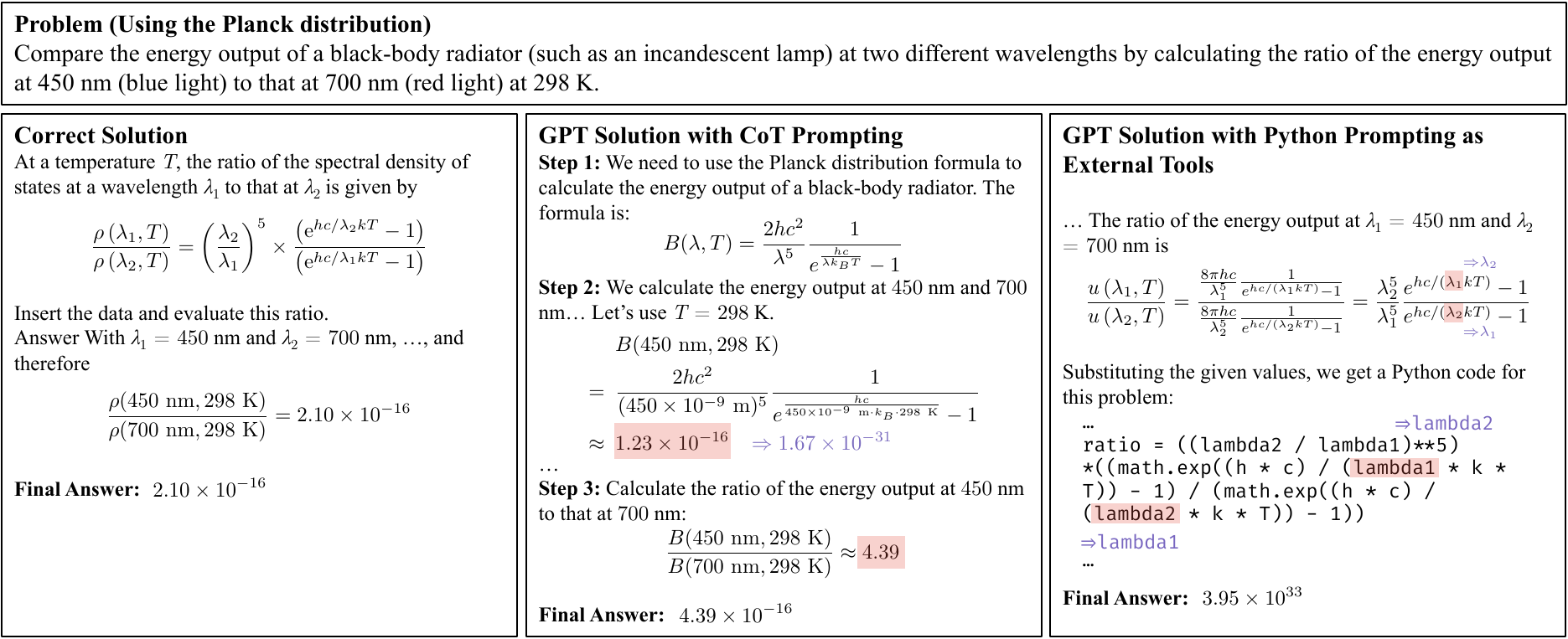}
    \caption{An example problem from \textit{Physical Chemistry} with solutions generated under two prompting strategies. GPT-4 with Chain-of-Thought (CoT) prompting shows calculation errors, while GPT-4 that prompts Python as external tools misunderstands mathematical equations. Errors are highlighted in {\color{sred}{red}} and the corrections are shown in {\color{spurple}{purple}}.}
    \label{fig:example}
    \vskip-0.8em
\end{figure*}

Recent advancements in Large Language Models (LLMs) have dramatically expanded the boundaries of artificial intelligence \cite{brown2020language, chatgpt, mmcot, llama, lamaadapter, lamaadapterv2, llava}.
They have demonstrated outstanding performance in many mathematical reasoning tasks that are typically considered challenging even for well-educated individuals \cite{wei2022chain,kojima2022large,codex,chen2022program,gao2022pal}.
Notably, GPT-4 achieves a remarkable score of 163 out of 170 on the GRE Quantitative Exam, placing it at the 80th percentile ranking \cite{gpt4}.

While the remarkable improvements in these benchmark performances might suggest that LLMs are capable of performing scientific reasoning tasks, we argue that this assertion might be overly optimistic due to the inherent limitations of current benchmarks.
Firstly, many existing benchmarks such as ScienceQA \cite{lu2022learn} and GSM8K \cite{cobbe2021training} only contain problems grounded in grade-level subjects.
Although other benchmarks like MATH \cite{hendrycks2021measuring} introduce high-school level questions, they primarily focus on math problems.
Secondly, recent works like MMLU \cite{hendrycks2020measuring}, AGIEval \cite{zhong2023agieval}, and JEEBench \cite{arora2023have}, despite introducing challenging problems that span a wide range of disciplines, only require basic computations---addition, subtraction, multiplication, and exponentiation---which do not adequately assess the depth of reasoning abilities of LLMs for solving scientific problems.
Lastly, most of these benchmarks only include textual problems, which omit problems that incorporate visual elements such as figures or diagrams.

In parallel to benchmark developments, many studies propose various prompting strategies aimed at enhancing the reasoning abilities of LLMs in scientific problem solving.
A notable example is the Chain-of-Thought (CoT) approach, which instructs LLMs to generate detailed, step-by-step solutions that prompt deeper problem thinking \cite{wei2022chain, wang2022self, zhou2022least, huang2022large}. Additionally, other strategies propose to enable LLMs to utilize external tools \cite{schick2023toolformer, lu2023chameleon} that improve the numerical computation capability.
However, even these strategic approaches, each with its specific strengths, struggle to fully address complex scientific problems.
Consider an example problem from college-level \textit{Physical Chemistry} \cite{atkins2014physical} that requires the use of the Planck distribution to derive certain quantities. As shown in \cref{fig:example}, LLMs with CoT prompts accurately generate the correct formula, but fail in the final numerical calculation.
As a remedy, when instructed to simultaneously generate a Python program for numerical computation and employ the CoT reasoning, the LLM misplaces $\lambda_1$ in the numerator rather than the denominator in the formula, illustrating a misunderstanding of mathematical relationships when employing external tools. This example highlights a crucial gap: even advanced LLMs struggle with complex scientific problem solving, necessitating a fine-grained analysis of the skills required for such complex tasks.

To mitigate these deficiencies, in this paper, we present a novel college-level \textbf{Sci}entific problem solving \textbf{Bench}mark, referred to as \ours.
\ours contains a carefully curated dataset of college-level scientific problems, including 869 problems collected from widely-used textbooks in college-level Chemistry, Physics, and Mathematics courses.
Distinct from existing benchmarks, all of the problems are open-ended, free-response questions that demand multi-step reasoning abilities, the understanding of scientific concepts, the retrieval of domain-specific knowledge (e.g., equations and theorems), and complex numeric computation capabilities (e.g., calculus or differential equations).
Besides that, our dataset includes a multimodal subset of 177 problems that incorporate visual elements (such as graphs and figures) as additional contexts, which enables of the evaluation of multimodal LLMs.
It is noted that \ours also includes step-by-step solutions for example problems, facilitating detailed error analysis.
To align our evaluation with real-world scenarios, we provide a separate, closed dataset that encompasses 103 problems from seven sets of midterm and final exams from collegiate Computer Science and Math courses.
To ensure the integrity of our evaluation, these datasets have been manually extracted from PDF documents and formatted into LaTeX documents, thereby minimizing the risk of their leakage in LLM training data.

Our evaluation includes a wide range of representative open-source and proprietary LLMs.
For unimodal, textual-based LLMs, we assess LLaMA-2, Mistral, Claude2, GPT-3.5, GPT-4, and their variants.
For multimodal vision-language models, we include GPT-4, InternLM-XComposer2, Qwen-VL, SPHINX-MoE, and LLaVA.
These models are tested using various prompting strategies, including CoT, zero-shot learning, and few-shot learning. We also prompt LLMs to utilize external scientific computing libraries in Python and Wolfram language.
The experimental results indicate that the complexity and difficulty of our dataset are sufficient to differentiate the performance levels of different LLMs.
Even with the strongest configuration---combining CoT prompting and the use of external tools---the best model achieves an average score of 43.22\% on the textual dataset, 13.8\% on the multimodal dataset, and 51.57\% on the closed exam dataset.
These results suggest a considerable potential for improvement in future LLMs.

\begin{table*}
\caption{Comparison of \ours with other benchmarks. ``Algebra'' refers to high-school level arithmetic computations; ``Calculus'' involves using integrals and differentials; ``Statistics'' focuses on applying statistical and probability concepts like bivariate distributions.}
\label{tab:compare}
	\SetTblrInner{rowsep=0.9pt}
	\begin{tblr}{
	colspec = {ccccccccccccc},
	row{1-2} = {bg=gray!25},
	row{even[3-14]} = {bg=gray!5}
	}
 \toprule
\SetCell[r=2]{c}Benchmark & \SetCell[c=3]{c}{Subject} & &  &  \SetCell[c=3]{c}{Calculation}& & & \SetCell[r=2]{c}\makecell{College\\Level}&\SetCell[r=2]{c}\makecell{Visual\\Contexts}&\SetCell[r=2]{c}\makecell{Detailed\\Solutions}&\SetCell[r=2]{c}\makecell{Free\\Response} \\
\cmidrule[lr]{2-4}\cmidrule[lr]{5-7}
  &  Math & Chemistry & Physics &  Algebra  & Calculus & Statistics  &  &  & & \\\midrule
ScienceQA\scite{lu2022learn}              & \checkmark &   \checkmark     &   \checkmark      &    \checkmark     &        &       &       &\checkmark&&       \\
IconQA\scite{lu2021iconqa} & \checkmark &   \checkmark     &   \checkmark      &      \checkmark     &        &       &       &\checkmark&&      \\
TabMWP \scite{lu2023dynamic}             & \checkmark &       &   &          \checkmark     &        &       &      &&\checkmark &   \checkmark    \\
GSM8K \scite{cobbe2021training}             & \checkmark &      &       &     \checkmark     &        &       &       &&\checkmark&  \checkmark     \\
MATH  \scite{hendrycks2021measuring}            & \checkmark &       & &      \checkmark     &        &       &      &&\checkmark&   \checkmark    \\
LILA \scite{mishra2022lila}& \checkmark &       & &      \checkmark     &        &       &      &&\checkmark&   \checkmark    \\

MMLU  \scite{hendrycks2020measuring} & \checkmark &    &        &           \checkmark     &        &       &       \checkmark  &&\checkmark &        \\
TheroemQA \scite{chen2023theoremqa}& \checkmark &       &        &    \checkmark   &   \checkmark     &       &    \checkmark   &&\checkmark &   \checkmark     \\ 
AGIEval \scite{zhong2023agieval}         & \checkmark &   \checkmark     &     \checkmark   &    \checkmark     &       &       &    \checkmark   &&&       \\
SciEval \scite{sun2023scieval}& \checkmark &   \checkmark     &   \checkmark          &    \checkmark     &        &       &      &&&       \\
JEEBench \scite{arora2023have}& \checkmark &   \checkmark     &   \checkmark          &    \checkmark     &        &       &      &&\checkmark&  \checkmark     \\
\midrule
\ours              & \checkmark &   \checkmark     &   \checkmark       &    \checkmark     &    \checkmark    &    \checkmark   &    \checkmark   &\checkmark&\checkmark&    \checkmark         \\
\bottomrule
\end{tblr}
\vskip-0.5em
\end{table*}

In order to gain a comprehensive understanding of the limitations of LLMs in scientific problem solving, we propose a novel self-refinement method to uncover the deficient skills in the solutions made by LLMs. Firstly, we compare the correct solutions with the solutions generated by LLMs and, with the assistance of human annotators, summarize ten essential skills requisite for successful scientific problem-solving. These skills include proficiency in domain knowledge, mathematical reasoning, numerical calculation abilities, and comprehension of common sense concepts. Subsequently, we employ an LLM-empowered self-critic approach to automatically classify the lacking skills in the solutions made by the benchmarked LLMs under each experiment configuration. Our analysis finds that (1) although CoT significantly improves the calculation ability, it is less effective in other aspects; (2) prompts with the use of external tools could potentially compromise other fundamental skills; (3) few-shot learning does not universally improve scientific problem-solving skills.

%% file: sections/related-work.tex
\section{Related Work}

Recently, many benchmarks have been proposed to assess the scientific problem-solving skills of LLMs, particularly in mathematical domains~\cite{lu2022survey,fu2023chain,lu2023dynamic,zhong2023agieval,mishra2022lila,chen2023theoremqa, guo2023indeed,hendrycks2020measuring, welleck2021naturalproofs}.
Notable works include GSM8K~\cite{cobbe2021training} including 8.5K grade school math word problems; LILA~\cite{mishra2022lila} which extends 20 datasets with task instructions and Python solutions; MATH~\cite{hendrycks2021measuring}, a challenging collection of 12.5K math problems from math competitions; TheroemQA~\cite{chen2023theoremqa}, focusing on theorem applications on problem solving; and MathVista~\cite{lu2023mathvista}, which evaluates the mathematical reasoning ability of LLMs in visual contexts.

To provide a more holistic evaluation, recent studies have expanded their scope to multiple disciplines:
ScienceQA~\cite{lu2022learn} introduces a multimodal question-answering dataset with accompanying lecture notes and explanatory annotations. \citet{taylor2022galactica} provide a set of scientific tasks, including LaTeX equation conversions, domain knowledge probes, citation prediction, and chemical question answering.
BIG-Bench~\cite{ghazal2013bigbench} offers a large-scale general-purpose test suite that requires 204 multiple-choice or exact-match tasks, and its extension BIG-Bench Hard~\cite{suzgun2022challenging} poses challenging CoT prompts.
SciEval~\cite{sun2023scieval} includes a mix of objective and subjective questions across multiple scientific fields to assess understanding, application, and research capabilities. JEEBench~\cite{arora2023have} incorporates pre-engineering-level scientific problems derived from college entrance exams.
AGIEval~\cite{zhong2023agieval} evaluates LLMs on human-centric standardized exams, such as college entrance exams and lawyer qualification tests.

Despite their extensive coverage across diverse disciplines, these datasets exhibit certain limitations. Sourced from lower educational level subjects, the majority of them focus on basic arithmetic operations rather than advanced mathematical computations.
Furthermore, most of these benchmarks are confined to textual-only problems, omitting problems with visual elements such as graphs or diagrams.
These drawbacks result in an incomplete assessment of the analytical and problem-solving skills required to tackle complex scientific problems.
In contrast, \ours focuses on college-level scientific problems across a broad spectrum of disciplines including Mathematics, Physics, and Chemistry. 
It emphasizes on a deep understanding of diverse scientific concepts, challenging LLMs to not only grasp these principles but also to efficiently retrieve and apply relevant knowledge.
Furthermore, it demands sophisticated numerical computation skills, including the execution of advanced mathematical operations such as calculus and differential equations, as well as the application of advanced statistical and probability theories.
Additionally, we include multimodal problems that necessitate the interpretation and integration of both textual and visual information.
A detailed comparison of \ours with some representative works is summarized in \cref{tab:compare}.

\begin{table*}
  \centering
  \caption{Summary of the textbook dataset. We report the number of total problems, percentage with detailed solutions, and percentage with visual elements in columns four to six respectively.}
	\SetTblrInner{rowsep=0.9pt}
	\begin{tblr}{
	colspec = {clcccc},
	row{1} = {bg=gray!25},
	row{5-8} = {bg=gray!10},
    column{3} = {font=\ttfamily},
	}
    \toprule
    Subject & \SetCell{c}Title & \SetCell{font=\normalfont}Acronym & \# Problems & \% Solutions & \% Visual \\
    \midrule
    \SetCell[r=3]{c}{Physics} & \textit{Fundamentals of Physics}  \scite{halliday2013fundamentals}   & fund  &    142   &   9.2\% & 43.0\% \\
          & \textit{Statistical Thermodynamics} \scite{engel2010thermodynamics}    & thermo &    83   &  20.5\% & 0.0\%\\
     & \textit{Classical Dynamics of Particles and Systems} \scite{thornton2021classical} &  class &  66     &  12.1\%   & 4.5\% \\
    \SetCell[r=4]{c}{Chemistry} & \textit{Quantum Chemistry} \scite{levine2009quantum}  & quan  &   41    &   19.5\% & 0.0\%\\
    & \textit{Quantum Chemistry} \scite{mcquarrie2008quantum} & chemmc &      47 &  19.1\% & 0.0\%\\
    & \textit{Physical Chemistry} \scite{atkins2014atkins} &  atkins &   122    &  13.9\% & 0.8\%\\
    & \textit{Physical Chemistry, Quanta, Matter, and Change} \scite{atkins2014physical} & matter &   59    &  16.9\% & 3.4\%\\
    \SetCell[r=3]{c} {Math} & \textit{Calculus: Early Transcendentals} \scite{stewart2012calculus} &  calc &    161   & 19.3\% & 67.7\%\\
          & \textit{Probability and Statistical Inference} \scite{hogg1977probability} & stat  &   93    & 21.5\% & 1.1\%\\
          & \textit{Elementary Differential Equations and Boundary Value Problems} \scite{boyce2021elementary} & diff  &      55 &9.1\%&0.0\%\\
          \bottomrule
    \end{tblr}
  \label{tab:stat}
  \vskip-0.5em
\end{table*}

While the aforementioned datasets focus on evaluating LLMs' performance on scientific problem solving tasks, another line of research aims to analyze the diverse capabilities of LLMs more comprehensively.
~\citet{liu2023evaluating} assess the reading abilities of LLMs using multiple-choice questions.
\citet{frieder2023mathematical} focus on evaluating the mathematical capabilities of LLMs, including those at the college level, but with topics such as functional analysis or topology that differ from those in \ours, such as differential equations and calculus.
\citet{bubeck2023sparks} explore the comprehensive abilities of GPT-4, but only use up to high-school level mathematical problems such as those in GSM8k~\cite{cobbe2021training} and MATH~\cite{hendrycks2021measuring}.
\citet{zhang2024sciglm} develop SciGLM, a scientific language model for collegiate-level problem reasoning, and evaluate its performance across multiple scientific datasets.
\citet{kabir2023answers} conduct a detailed manual analysis for LLMs. They also provide human-annotated qualitative analysis to assess the capabilities of the models.
However, relying on human labor for direct solution analysis can be costly.
Our evaluation protocol, based on predefined fundamental problem solving skills, enables automated classification of deficient skills for each incorrectly answered question. This approach enables an affordable, large-scale qualitative analysis of model solutions.

%% file: sections/datasets.tex
\section{The \ours Dataset}
\label{data}
To evaluate the capabilities and analyze the limitations of Large Language Models (LLMs) to solve scientific computing problems, we collect a new dataset consisting of college-level textbooks and course exams in a variety of domains.
This section details the dataset construction process.

\textbf{Data selection criteria.} 
Our dataset aims to improve the previous benchmarks by including more challenging problems.
Specifically, the selected dataset should fulfill the following requirements:
\begin{itemize}
    \item  \textbf{Inclusion of college-level problems.} The chosen problems demand a solid understanding of domain-specific knowledge, adept calculation skills, and the ability to perform complex numerical computations.
    \item \textbf{Inclusion of detailed solutions.} To facilitate a thorough analysis of the limitations of LLMs, detailed solutions should be provided as well, which could facilitate a finer-grained examination of the capacity of LLMs to handle complex problem-solving tasks.
    \item \textbf{Inclusion of visual elements.} In the real world, many scientific problems require the interpretation and integration of both textual and visual information. The included problems should thus contain visual elements (such as figures) in the contexts.
    \item \textbf{Inaccessibility in text formats.} To ensure an unbiased evaluation, questions should not be readily accessible online and cannot be easily extracted or transformed into text. This aims to mitigate any potential information leakage from the exposure of LLMs to pre-existing online question banks, such as those found in standardized tests like the SAT exams.
    \item \textbf{Assessment of advanced problem-solving capabilities.}
    The problems to benchmark should not be confined to basic arithmetic operations like addition and multiplication. Rather, they should enable evaluating the capability of LLMs in performing advanced computations such as calculus and differential equations. 
\end{itemize}

Accordingly, to construct the dataset, we select ten textbooks from three scientific fields Physics, Chemistry, and Mathematics that have been extensively used in college courses.
We summarize the statistics of this textbook dataset in \cref{tab:stat} and we use acronyms to refer to each textbook throughout the paper for brevity.
Furthermore, in order to simulate real-world evaluation, we compile a closed set of exam questions from college courses from Computer Science and Math departments, including \textit{Data Mining}, \textit{Machine Learning}, and \textit{Differential Equations}.
This subset is less likely to be in LLM training data, making it an effective tool for LLM evaluation.
Detailed statistics of these exam problems are summarized in \cref{tab:new-exam}.
We refer readers to \cref{textbook} for details on these textbooks and exams.

To reduce the likelihood of correct answers being merely guessed from candidates, we choose to mainly include questions with more challenging, free-response answers, rather than multiple-choice questions in previous works \cite{lu2021inter, lu2022learn,  chen2023theoremqa}.
In order to facilitate standardized and automated evaluation, we focus on answers that only contain single numerical numbers to avoid ambiguity for the textbook dataset.
Further, we convert the answer to floating-point numbers rounded to three decimal places. For example, the answer $\frac{\sqrt{2}}{\pi}$ will be converted to the decimal representation of $0.450$. We also treat scientific notation as a unit to avoid overflow issues. For example, if the answer is $2.2\times 10^{-31}~\mathrm{m}$, we take $2.2$ as the final answer and $10^{-31}~\mathrm{m}$ as the unit.

\textbf{Data preprocessing.}
We collect each problem from the original textbooks in PDF documents and manually process them into LaTeX documents using an OCR tool \href{https://mathpix.com/}{Mathpix}. The data is manually collected by human annotators using a web-based annotation tool \cite{lu2021inter}, whose user interface is shown in \cref{appendix:data}.
All problems are carefully verified by human annotators to ensure that LaTeX documents can be compiled without any syntax errors. For reference, we also provide the original numbers in textbooks.
For every problem, we provide the answer in two forms: the numerical value and the corresponding LaTeX expression with mathematical notations retained (e.g., $0.450$ and $\frac{\sqrt{2}}{\pi}$); the unit of each answer is saved as a separate attribute. %
The detailed step-by-step solutions are also provided in LaTeX.
For problems having multiple answers, we either keep only the first subproblem and discard the remaining subproblems or convert each subproblem into a separate problem.

\begin{table*}[t]
\caption{Experimental results in terms of accuracy (\%) on the textbook dataset. The best performing score is highlighted in \textbf{bold} and second-best is \ul{underlined}. The average score is weighted by the number of problems in each textbook.}
\label{res2}
\centering
\resizebox{0.85\textwidth}{!}{
\SetTblrInner{rowsep=0.9pt}
\begin{tblr}{
colspec = {cccccccccccc},
row{2} = {font=\ttfamily},
row{1-2} = {bg=gray!25},
row{3,11,19,27} = {font=\itshape},
hline{4,12,20,28} = {dashed},
row{11-18,27-34} = {bg=gray!10}
}
\toprule
\SetCell[r=2]{c}\normalfont{Model} & \SetCell[c=4]{c}{Chemistry}  & & & &\SetCell[c=3]{c}{Physics} & & & \SetCell[c=3]{c}{Math} & & & \SetCell[r=2]{c}{Avg.}  \\
\cmidrule[lr]{2-5}\cmidrule[lr]{6-8}\cmidrule[lr]{9-11}
 & atkins & chemmc & quan & matter & fund  & class  & thermo & diff  & stat  & calc &         \\\midrule
\SetCell[c=12]{c}{Zero-Shot Learning} \\
    LLaMA-2-7B & 0.00  & 0.00  & 0.00  & 0.00  & 1.37  & 0.00  & 0.00  & 2.00  & 5.33  & 0.00  & 1.03 \\
    LLaMA-2-70B & 1.87  & 2.56  & 0.00  & 0.00  & 1.40  & 0.00  & 0.00  & 0.00  & 10.70 & 4.76  & 2.41 \\
    Mistral-7B & 9.35 & 5.13 & 8.82  & 4.08  & 5.48  & 2.13  & 0.00  & 4.00  & 12.00 & 2.38 & 6.23 \\
    Claude2 & 15.00 & 12.83 & 14.71 & 10.20 & 12.33 & 6.40  & 9.00  & 4.00  & 38.70 & 16.70 & 14.94 \\
    GPT-3.5-Turbo & 4.67  & 20.51 & 8.82  & 2.04  & 10.96 & 2.13  & 2.94  & 6.00  & 28.00 & 9.30  & 9.59 \\
    GPT-4 & \ul{45.79} & \ul{28.21} & \ul{26.47} & \ul{22.45} & \ul{23.29} & \textbf{25.53}  & \ul{17.91} & \ul{32.00} & \ul{49.33} & \textbf{54.76} & \ul{33.79} \\
    GPT-4-Turbo & \textbf{57.01} & \textbf{41.03} & \textbf{35.29} & \textbf{26.53} & \textbf{24.66} & \ul{21.28} & \textbf{26.87} & \textbf{46.00} & \textbf{61.33} & \ul{52.38} & \textbf{40.99} \\\midrule
\SetCell[c=12]{c}{Zero-Shot Learning + CoT Prompting} \\
    LLaMA-2-7B & 0.00  &  2.56 &  0.00 & 0.00 &  0.00& 0.00  &   0.00& 0.00  & 4.00  &  0.00& 0.67 \\
    LLaMA-2-70B & 0.93 &  2.56 &  0.00 & 0.00 &  0.00& 0.00  & 1.49 & 0.00  & 10.70  &  0.00 &  1.89 \\
    Mistral-7B & 6.54  & 5.13 & 2.94 & 0.00 & 0.00 & 2.12  & 1.49  & 6.00  & 10.67  & 9.52  & 4.63 \\
    Claude2 & 20.56 &  15.38&  8.82 & 4.08 &  8.23&4.26& 5.97  &   6.00 &   36.00  &  14.29 &13.89 \\
    GPT-3.5-Turbo & 6.54 &  23.08 &  2.94 & 10.20 &  12.33& 2.12  & 5.97 & 12.00  &   33.33  &  9.30&   12.17 \\
    GPT-4 & \ul{28.04}     &    \textbf{43.59}   &    \ul{14.71}  &  \ul{20.41} &   \ul{21.92}    & \ul{19.15} &    \ul{17.91}    &    \ul{22.00}   &   \ul{50.67} &  \ul{42.86}&   \ul{28.52} \\
    GPT-4-Turbo & \textbf{60.75} & \ul{35.90} & \textbf{29.41} & \textbf{28.57} & \textbf{30.14} & \textbf{31.91} & \textbf{25.37} & \textbf{38.00} & \textbf{64.00} & \textbf{54.76} & \textbf{42.37} \\\midrule
\SetCell[c=12]{c}{Few-Shot Learning + CoT Prompting} \\
    LLaMA-2-7B & 1.87  & 5.13  & 2.94  & 0.00  & 5.48  & 0.00  & 0.00  & 0.00  & 12.00 & 7.14  & 3.60 \\
    LLaMA-2-70B & 13.10 & 12.83 & 14.71 & 4.08  & 12.33 & 0.00  & 0.00  & 0.00  & 13.30 & 9.52  & 8.40 \\
    Mistral-7B & 6.54  & 10.26 & 2.94  & 2.04  & 2.74  & 2.13  & 4.48 & 4.00  & 14.67 & 9.52 & 6.17 \\
    Claude2 & 15.89 & 25.64 & 14.65 & 6.12  & 9.59  & 6.38  & 10.45 & 8.00  & 33.33 & 19.05 & 15.26 \\
    GPT-3.5-Turbo & 8.41  & 20.51 & 8.82  & 6.12  & 10.96 & 2.12  & 1.49  & 10.00 & 38.67 & 6.98  & 11.99 \\
    GPT-4 & \ul{41.12} & \ul{33.33} & \ul{17.65} & \ul{16.33} & \ul{17.81} & \ul{17.02} & \ul{20.90} & \ul{30.00} & \ul{49.33} & \ul{45.24} & \ul{30.36} \\
    GPT-4-Turbo & \textbf{59.81} & \textbf{35.90} & \textbf{26.47} & \textbf{18.37} & \textbf{23.29} & \textbf{19.15} & \textbf{32.84} & \textbf{32.00} & \textbf{65.33} & \textbf{50.00} & \textbf{39.45} \\\midrule
\SetCell[c=12]{c}{Few-Shot Learning + Python} \\
    LLaMA-2-7B & 0.93  & 2.56  & 0.00  & 0.00  & 0.00  & 0.00  & 0.00  & 0.00  & 6.67  & 0.00  & 1.20 \\
    LLaMA-2-70B & 0.93  & 7.69  & 2.94  & 0.00  & 9.59  & 0.00  & 1.49  & 0.00  & 17.30 & 9.52  & 5.14 \\
    Mistral-7B & 4.67  & 0.00  & 5.88  & 2.04  & 2.74  & 2.13 & 0.00  & 4.00  & 17.33  & 11.90  & 5.32 \\
    Claude2 & 6.54  & 12.82 & 14.71 & 4.08  & 17.81 & 8.51  & 5.97  & 20.00 & 40.00 & 16.67 & 14.92 \\
    GPT-3.5-Turbo & 13.08 & \ul{33.33} & 8.82  & 16.33 & 26.01 & 4.26  & 7.46  & 16.00 & \ul{44.00} & 26.19 & 19.91 \\
    GPT-4 & \textbf{57.01} & \textbf{38.46} & \textbf{44.12} & \textbf{34.69} & \textbf{28.77} & \textbf{23.40} & \textbf{34.33} & \textbf{44.00} & \textbf{68.00} & \textbf{38.10} & \textbf{43.22} \\
    GPT-4-Turbo & \ul{32.71} & \ul{33.33} & \ul{17.65} & \ul{26.53} & \ul{27.40} & \ul{12.76} & \ul{16.42} & \ul{34.00} & 42.67 & \ul{30.95}&\ul{28.47 }\\
    \bottomrule
\end{tblr}
}
\end{table*}

%% file: sections/experiments.tex
\section{Experiments}
This section presents the experiments to assess the capabilities of LLMs in scientific problem-solving. We first describe our experimental setup. Subsequently, we evaluate unimodal LLMs on the textbook dataset. Following this, we include additional experiments on the multimodal subset and the closed exam subset, as well as comparisons with other numerical computational tools.

\subsection{Experiment Setup}
We evaluate the textbook dataset on seven unimodal LLMs, which include four proprietary models: Claude2 (\texttt{claude2}) \cite{claude}, GPT-3.5-Turbo (\texttt{gpt}\texttt{-3.5}\texttt{-turbo}) \cite{chatgpt}, GPT-4 (\texttt{gpt-4}), GPT-4-Turbo (\texttt{gpt-4-turbo}) \cite{gpt4}, along with three open-source models: LLaMA-2-7B (\texttt{llama-2-7b-chat}), LLaMA-2-70B (\texttt{llama-2-70b-chat})\cite{touvron2023llama}, and Mistral-7B (\texttt{mistral-7b-instruct})\cite{jiang2023mistral}.

We consider two prompting strategies, including the Chain-of-Thought (CoT) prompting and prompting to use external tools.
\label{exp}
\label{sec:exp_set}
\begin{itemize}
    \item \textbf{Zero-shot and few-shot learning.}
In the zero-shot learning setting, models are not provided with any prior examples, which evaluates their inherent problem-solving capabilities with background knowledge and reasoning abilities. In the few-shot setting, a few examples are given to the models before the test example. This aims to assess their capability to learn new information from the demonstrations and incorporate it into their problem-solving processes.

    \item \textbf{Prompting-based approaches.}
For our experiments, all settings begin with a system prompt that describes the types and categories of questions. Additionally, we utilize a CoT prompting strategy in zero- and few-shot settings.

    \item \textbf{Tool-augmented approaches.}
Given that LLMs are limited in acquiring exact knowledge and performing precise calculations, some recent approaches, such as PAL \cite{gao2022pal} and PoT \cite{chen2022program} 
explore utilizing external tools such as the Python interpreter for program synthesis to enhance the capabilities of solving complex reasoning tasks. In line with these approaches and acknowledging the limitations of LLMs in performing precise calculations, we also include a setting that prompts the model to convert its solution steps in natural language into Python code, aiming to achieve more accurate results for certain computation steps. This tool-augmented approach can only be tested in the few-shot learning setting. We manually construct Python programs that produce the correct answer.

\end{itemize}

\textbf{Implementation details.} 
We set temperature to zero for all models to reduce the randomness of the predictions.
Few-shot examples, including solutions, are randomly selected from problems within each textbook. When external tools are used, we add a code snippet that translates the solution into specific programming languages in all few-shot examples. The code snippets are verified by human annotators that will produce the correct output.
In terms of evaluation metrics, we compare the model outputs with the correct answers, allowing a relative tolerance of 5\%.
In particular to the exam dataset, the model solutions are graded using the rubrics provided by the instructors.
Readers may refer to \cref{appendix:details} for all prompts and the implementation details for utilizing external tools.

\subsection{Results and Analysis}
\label{exp_res}

We report the model performance in terms of accuracy score for each textbook and an average score over all problems.
The results of all LLMs in various settings on the textbook and the exam dataset are summarized in \cref{res2,exam_result} respectively.
We have the following observations.

\begin{itemize}
    \item \textbf{Observation 1. \ours is complex enough to differentiate among LLMs.}
    Our results show that open-source models such as LLaMA-2 and Mistral are consistently outperformed by their proprietary counterparts across all settings within the textbook dataset. Notably, GPT-4 and GPT-4-Turbo lead in performance by a significant margin. For example, GPT-4-Turbo outperforms Mistral-7B by 34.76\% in the zero-shot setting.
    Additionally, within both LLaMA and GPT series, we observe a clear correlation between increased model capacity (i.e., larger parameter sizes) and improved performance.
    Therefore, the complexity of \ours is able to differentiate the performance among different LLMs.
    \item \textbf{Observation 2. \ours highlights varied efficacy of prompting strategies across LLMs.}
    Our findings suggest that the effectiveness of employing prompting strategies or external computational tools varies significantly among different LLMs.
    As shown in the table, LLaMA-2-70B shows a marked improvement in the few-shot setting over the zero-shot setting, increasing from 2.41\% to 8.40\%.
    Similarly, the performance of GPT-4 is significantly improved when incorporating external tools, with an increase from 30.36\% to 43.22\%.
    Meanwhile, the up-to-date model GPT-4-Turbo exhibits superior performance in zero-shot learning settings.
    However, despite its advanced capabilities demonstrated by its outstanding zero-shot learning performance, it falls short compared to GPT-4 in few-shot learning when leveraging Python for numerical computation. This suggests a potential reduction in its program understanding capabilities.
    In summary, such findings illustrate \ours can reveal the nuanced differences in the ability of LLMs to utilize prompting strategies and external tools effectively.

\end{itemize}

\begin{figure}[t]
	\centering
	\includegraphics[width=\linewidth]{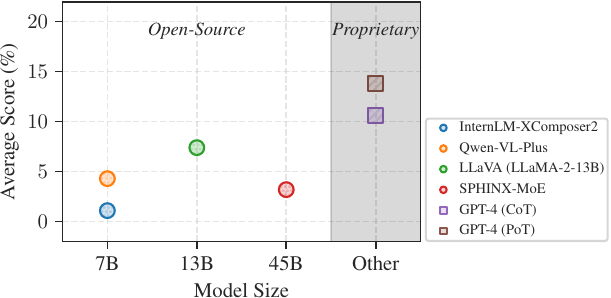}
	\caption{Performance of LLMs on the multimodal subset. GPT-4 models are augmented with image captions and OCR text.}
	\label{fig:vision}
 \vskip-0.5em
\end{figure}

\begin{figure*}
    \centering
    \includegraphics[width=\linewidth]{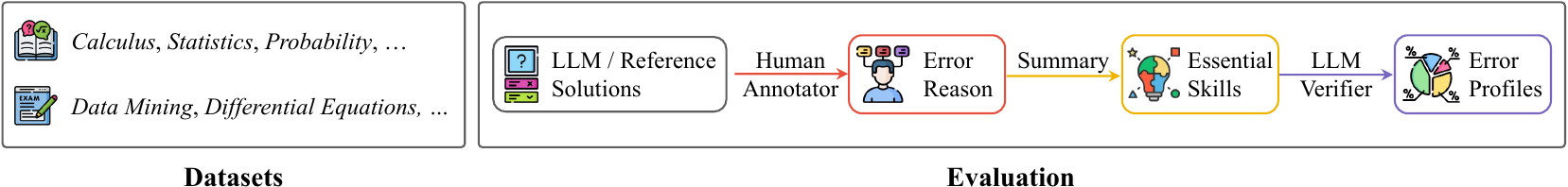}
	\vskip-0.5em
    \caption{Pipeline of the evaluation protocol. 
    The evaluation protocol involves analyzing both LLMs and reference (correct) solutions with the assistance of human annotators to identify error reasons. These reasons are then summarized into ten essential scientific problem-solving skills in which LLM may face challenges. Subsequently, a LLM verifier is employed to automatically attribute each incorrectly answered problem to a lack of a specific skill. The resulting error profiles enable the interpretation of the improved skills by certain prompting strategies and the direct comparison of various strategies.
    }
    \label{fig:pipeline}
\end{figure*}

\subsection{Additional Experiments}
\textbf{Evaluation on the multimodal subset.}
We evaluate two categories of models on problems with visual contexts: (1) GPT-4 \cite{gpt4} augmented with image captions from Multimodal Bard \cite{googlebard} and OCR texts from EasyOCR \cite{easyocr} and (2) open-source Large Multimodal Models (LMMs): InternLM-XComposer2-VL \cite{internlmxcomposer2}, Qwen-VL-Plus \cite{Qwen-VL}, SPHINX-MoE \cite{lin2023sphinx}, and LLaVA-LLaMA-2-13B \cite{liu2023llava}.
For GPT-4, we explore two prompting strategies: Chain-of-Thought (CoT) \cite{wei2022chain} and Program-of-Thoughts (PoT) \cite{chen2022program}.
The results presented in \cref{fig:vision} reveal that proprietary models augmented with image captions and OCR-detected text, significantly outperform their open-source counterparts. GPT-4 (PoT) that combines programming capabilities achieves an accuracy of 13.8\%, markedly higher than 7.4\% obtained by the best open model LLaVA-LLaMA-2-13B. This demonstrates the substantial potential for LLMs to effectively utilize visual contexts in scientific problem solving.

\textbf{Evaluation on the exam subset.}
To mirror real-world testing conditions with no few-shot examples provided, we evaluate GPT-3.5, GPT-4, Claude, LLaMA-2-7B, and LLaMA-2-70B on the closed exam dataset under zero-shot and zero-shot CoT settings. The experiment results summarized in \cref{exam_result} indicate a notable performance advantage of GPT-4, which achieves an averaged score of 57.54\%. However, we note that their performance remains significantly lower than human benchmarking. For instance, in the Data Mining course, GPT-4 scores 64.44\% and 42.67\% in the midterm and final exams, lower than the average student scores of 80.18\% and 72.71\%, respectively, as reported by the course instructor. The results once again underline the challenging nature of our dataset.

\textbf{Comparison with other scientific computing tools.}
We further utilize another famous scientific computing library \href{https://www.wolfram.com/language/}{Wolfram Language} as the external tool and conduct experiments using GPT-3.5, Claude, LLaMA-2-7B, and LLaMA-2-70B. The experiment results reported in \cref{fig:wolfram} show that utilizing Wolfram Language does not help few-shot learning and even results in a deteriorated performance, with a decrease of 6.70\% compared to the CoT prompting for Claude2, and a decrease of 6.17\% for LLaMA-2-70B. A plausible explanation is the introduction of syntax errors when translating solution steps into the Wolfram Language, which could be a potential direction for improvement.
For a detailed error analysis, readers are directed to \cref{appendix:wolfram}.

%% file: sections/analysis.tex
\begin{figure*}
\includegraphics[width=\linewidth]{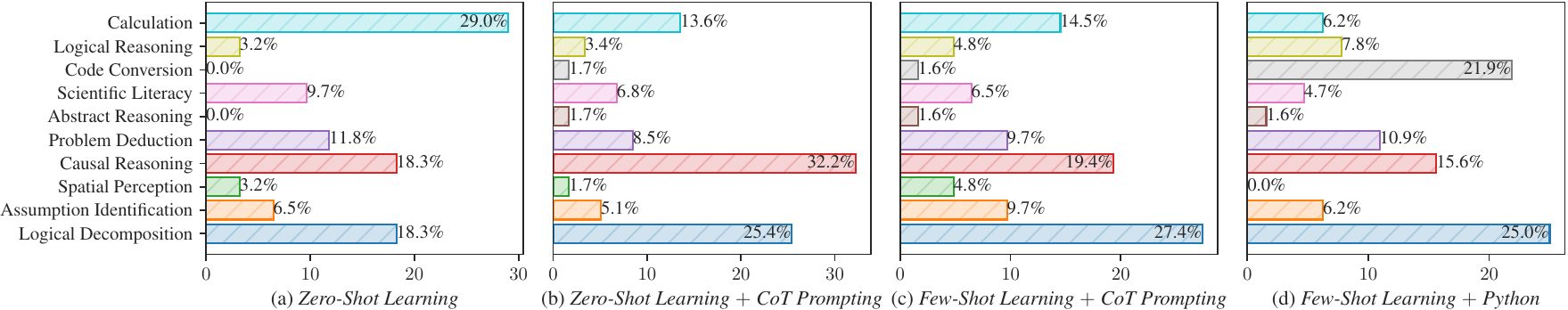}
\caption{Error profiles of GPT-3.5 on the textbook dataset under four settings, which reveal the distribution of their deficiencies in ten essential problem-solving abilities.}
\label{reason}
\vskip-0.5em
\end{figure*}

\section{Error Analysis of Prompting Strategies}
\label{sec_ana}

Considering the substantial advancements of current LLMs, an in-depth analysis of the particular skills that are either enhanced or limited under certain settings becomes imperative. Previous works have relied on human labor to annotate error reasons into different categories, which is both expensive and time-consuming\cite{zhong2023agieval}. In this section, we present an evaluation protocol that automates the classification of error reasons into deficient skills. This time-efficient approach enables large-scale analyses in future research.

In order to quantify the impact of each setting on scientific problem-solving, we first define an essential skill set that is required by solving scientific problems.
Then, an LLM verifier is employed to automatically classify each incorrectly solved problem based on the absence of a specific skill from the essential skill set. This approach generates error profiles, showcasing a direct comparison of different strategies. 
This evaluation protocol is summarized in \cref{fig:pipeline}.

Firstly, we analyze the incorrect solutions made by GPT-3.5 for problems that provide detailed solutions.
We hire two college students, who are highly familiar with the problems in our datasets, to annotate the source of the error for each problem, indicating the specific line where the model makes a mistake and why. From 112 such error annotations and with the assistance of GPT-4, we distill these errors into ten essential skills that GPT-3.5 might lack:
\begin{itemize}
    \item \textbf{Logical decomposition and analysis skills.} This ability involves decomposing the problem into smaller, manageable parts, and understanding the relationships between these parts.
    \item \textbf{Assumption identification.} This skill involves the ability to recognize relevant and necessary assumptions in the problem.
    \item \textbf{Spatial perception.} This is important for understanding problems in areas such as Physics and Chemistry, where models need to visualize molecules, forces, fields, etc.
    \item \textbf{Causal reasoning.} This is the ability to understand cause and effect relationships.
    \item \textbf{Problem deduction skills.} This pertains to the ability to infer and deduce potential solutions or underlying principles from the given information in a problem.
    \item \textbf{Abstract reasoning.} This skill involves the ability to understand complex concepts that cannot be perceived physically, and to recognize patterns or relationships beyond concrete examples.
    \item \textbf{Scientific literacy.}  This skill involves a comprehensive understanding of key scientific principles, terminology, and methodologies across a range of disciplines.
    \item \textbf{Code conversion skills.} This involves the ability to accurately translate solution steps into different programming languages, like Python or Wolfram Language.
    \item \textbf{Logical reasoning.} This is the ability to make a reasoned argument and to identify fallacies or inconsistencies in an argument or set of data.
    \item \textbf{Calculation skills.} This involves the ability to accurately carry out mathematical operations and computations.
\end{itemize}

After identifying this essential skill set, we assess the performance of the LLMs under different settings to discern the specific problem-solving skills they lack.
Given the high cost of human annotations required to attribute the cause of incorrect solutions to specific skill deficiencies, we propose a novel self-critique protocol: we design a specific prompt that outlines these abilities, and employ another LLM to serve as a classifier and determine whether a specific error results from the lack of a particular problem-solving skill. Finally, we ask human annotators to scrutinize the classification results, which results in approximately 20\% of incorrectly classified skills being discarded.
To be specific, we utilize a GPT-3.5 model as the verifier to determine the reason behind each error and pinpoint the missing skill. The details regarding the specific prompts used are provided in \cref{prompt}. This verification process is conducted for four settings, with results represented in bar charts (\cref{reason}). Additional examples of the evaluation protocol are elaborated in \cref{reasoning}.

Our findings suggest that \textbf{there is a lack of a universally effective setting: each configuration only enhances some specific abilities and occasionally even hurts other skills that the original model possesses.}
First, CoT prompting significantly improves calculation skills in the zero-shot scenario, with  13.6\% error rates caused by calculation ability, considerably lower than the 29.0\% error rate of the vanilla zero-shot baseline. However, CoT shows limitations in improving other skills, with 32.2\% and 25.4\%  error rates in casual ability and logical decomposition ability in the zero-shot CoT setting, respectively, compared to 18.3\% and 18.3\% in the zero-shot setting. This contradicts previous claims about universal skill enhancement through zero-shot CoT and carefully-designed few-shot CoT prompts\cite{wei2022chain}. An example in \cref{appendix:eg_error} shows that the zero-shot learning setting without CoT has generated the correct formula but fails in the calculation steps. In this case, CoT prompting is even unable to use the correct formula as it misinterprets the specific conditions (non-necessity) in the problem. Second, the use of external tools significantly reduces calculation errors compared to the few-shot Cot setting, with a notable decrease from 14.5\% to 6.2\%. However, the use of external tools can weaken other skills, particularly the code conversion skills, i.e., generating the correct programs for the solution. 
Third, few-shot learning does not universally improve scientific problem-solving skills, as indicated in the comparison between zero-shot and few-shot CoT settings. The improvement in one skill is offset by the shortcomings in others: although the few-shot CoT setting results in a reduction of 12.8\% in errors related to causal reasoning, it also leads to an increase in errors associated with other skills, such as logical decomposition.

%% file: sections/conclusion.tex
\section{Conclusion}
This paper presents \ours, a college-level benchmark that includes scientific problems from Mathematics, Physics, and Chemistry, as well as exam questions in Computer Science and Mathematics. Our comprehensive evaluation includes a diverse array of Large Language Models (LLMs), spanning both open-source and proprietary models, including unimodal as well as multimodal settings, and employing a variety of prompting strategies.
The evaluation protocol we employ serves as a framework for evaluating advanced problem-solving skills of LLMs in scientific domains. The findings of this study highlight that while large language models (LLMs) exhibit impressive performance on introductory mathematical benchmarks, their mastery of problem solving ability remains weak. These findings underscore the limitations of current LLMs in achieving satisfactory performance, even with the assistance of various tools. We envision that the \ours benchmark dataset and evaluation protocol presented in this paper could lay a foundation for future research and enable advancements in understanding and enhancing problem-solving capabilities of LLMs. 

%% file: sections/reproducibility.tex
\section*{Reproducibility Statement}
To foster reproducible research, we include all dataset processing and experiment details of \ours. We detail data processing in \cref{data} and provide the UI design of data collection in \cref{appendix:data}. We include all experiment details with LLM prompts in \cref{appendix:details}. Finally, we make our dataset and code publicly available at \href{https://github.com/mandyyyyii/scibench}{this repository}.

%% file: sections/ethical.tex
\section*{Ethical Statement}
The questions of \ours are sourced from science textbooks and exams. We conduct a manual examination of our dataset to ensure the absence of potential sensitive background or ethical concerns. The inclusion of exam questions has been authorized by the instructors of the respective courses. 

The purpose of the textbook dataset is solely for academic use. Its collection adheres to the \emph{Fair Use Law} in the US, where only a certain number of questions from each textbook are selected, ensuring that only a small portion of the textbook is utilized.

%% file: sections/impact.tex
\section*{Impact Statement}
The introduction of \ours represents a significant advancement in the evaluation of Large Language Models (LLMs) for scientific problem-solving tasks. By focusing on collegiate-level problems in mathematics, chemistry, and physics, \ours addresses a critical gap in existing benchmarks, which have primarily focused on high-school subjects and basic algebraic operations. This development underscores the necessity of developing specialized benchmarks that challenge LLMs with higher complexity problems, thereby pushing the boundaries of the capabilities of LLMs in academic and research settings.

While the current scope of \ours encompasses a select group of scientific disciplines, the potential for future extensions is vast. Incorporating additional subjects such as biology, computer science, and engineering could provide a more comprehensive understanding of LLM capabilities across a broader spectrum of scientific knowledge. Moreover, extending the benchmark to social sciences, humanities, and other human-centric domains would be equally beneficial, as these areas often involve nuanced reasoning and interpretation of complex social dynamics and ethical considerations, posing unique challenges that could further enhance the versatility and applicability of LLMs.

%% file: sections/appendix.tex
\clearpage
\appendix
\onecolumn
\appendixprefix
\begin{center}
	{\Large \textbf{Supplementary Material for \ours}}
\end{center}

\newenvironment{titledframe}[1]
  {\mdfsetup{
    frametitle={\colorbox{white}{\space#1\space}},
    innertopmargin=10pt,
    frametitleaboveskip=-\ht\strutbox,
    frametitlealignment=\center
    }
  \begin{mdframed}%
  }
  {\end{mdframed}}

\startcontents[sections]
\printcontents[sections]{l}{1}{\setcounter{tocdepth}{2}}

\section{The Textbook Dataset}
\label{textbook}
\subsection{Textbook Sources}
\begin{itemize}
    \item \textsc{Physical Chemistry \cite{atkins2014atkins}} (\texttt{atkins}) provides an exploration of equilibrium, structure, and reactions, integrating contemporary techniques like nanoscience, spectroscopy, and computational chemistry. 
    \item \textsc{Quantum Chemistry \cite{mcquarrie2008quantum}} (\texttt{chemmc}) meticulously covers Quantum Mechanics, from foundational principles like blackbody radiation and Heisenberg's Uncertainty Principle to complex topics such as Schrödinger equation, quantum mechanical operators, and the application of quantum mechanics in chemical bonding.
    \item \textsc{Quantum Chemistry \cite{levine2009quantum}} (\texttt{quan}) explores quantum chemistry, providing a detailed understanding of the Schrödinger equation, particle behavior in various scenarios, quantum mechanics operators, and other foundational quantum principles. It delves into specific applications like the electronic structure of diatomic and polyatomic molecules, variation methods, perturbation theory, electron spin and its implications in quantum mechanics, as well as various computational methods for molecular quantum mechanics.
    \item \textsc{Physical Chemistry, Quanta, Matter, and Change \cite{atkins2014physical}} (\texttt{matter}) combines physics and mathematics, beginning with basics like differentiation and integration, advancing through quantum mechanics and atomic structure, then exploring thermodynamics, molecular motion, and chemical kinetics. Each section is supplemented with mathematical concepts such as differential equations, vectors, and probability theory.
    \item \textsc{Classical Dynamics Of Partical and Systems \cite{thornton2021classical} }(\texttt{class}) initiates with an exploration of fundamental mathematical concepts, discussing scalars, vectors, matrix operations, coordinate transformations, differentiation, and integration of vectors, using these constructs to illustrate concepts like velocity, acceleration, and angular velocity. It then transitions into the realm of Newtonian mechanics, detailing Newton's laws, frames of reference, and the equation of motion for a single particle.
    \item \textsc{Thermodynamics, Statistical Thermodynamics, and Kinetics \cite{engel2010thermodynamics}} (\texttt{thermo}) navigates through thermodynamics' principles, from fundamental concepts to complex laws, further discussing real and ideal gases, solutions, electrochemical cells, and statistical thermodynamics. It concludes with an examination of the kinetic theory of gases, transport phenomena, and chemical kinetics.
    \item \textsc{Fundamentals of Physics \cite{halliday2013fundamentals}} (\texttt{fund}) covers undergraduate physics topics, ranging from fundamental concepts like motion and energy to more advanced areas such as quantum physics and nuclear physics.
    \item \textsc{Elementary Differential Equations and Boundary
    Value Problems \cite{boyce2021elementary}} (\texttt{diff}) provides a detailed exploration of differential equations, progressing from basic mathematical models to advanced topics like the Laplace Transform, linear systems, numerical methods, and Fourier series. It culminates with a deep dive into nonlinear equations, partial differential equations, and boundary value problems.
    \item \textsc{Probability and Statistical Inference \cite{hogg1977probability}} (\texttt{stat}) covers probability and statistics, including fundamental concepts, discrete and continuous distributions, bivariate distributions, functions of random variables, and estimation techniques.
    \item \textsc{Calculus: Early Transcendentals \cite{stewart2012calculus}} (\texttt{calculus}) begins with diagnostic tests in foundational topics, and explores functions from multiple perspectives. It comprehensively covers calculus concepts from limits to three-dimensional analytic geometry, incorporating applications in various fields.
\end{itemize}

\subsection{Textbook Examples}
The textbook examples are provided in \cref{appendix:subj}. The examples from the multimodal subset are provided in \cref{img_eg0,image_eg1,img_eg2,img_eg3}.
\input{tables/subj_eg}

\input{tables/img_eg0}
\input{tables/img_eg1}
\input{tables/img_eg2}
\input{tables/img_eg3}

\subsection{UI Design of the Labeling Tool}
We employed a team of seven individuals to gather data from textbooks using an annotation tool. Each individual was responsible for one to two books, encompassing approximately 100 examples. The user interface of the annotation tool is depicted in \cref{appendix:ui}. For subsequent verification, we preserved images of problems and their corresponding answers. To ensure clarity in future references, we have maintained the original sequence of problems as they appear in the textbooks.
\label{appendix:data}

\begin{figure*}
\centering
\fbox{\includegraphics[width=\textwidth]{figures/ui.pdf}}
\caption{The UI design of data annotation.}
\label{appendix:ui}
\end{figure*}

\section{The Exam Dataset}

The exam dataset is drawn from the following sources:
\begin{itemize}
    \item \textsc{Introduction to Data Mining} provides an introductory survey of data mining, which involves the automatic discovery of patterns, associations, changes, and anomalies in large databases. It explores various application areas of data mining, including bioinformatics, e-commerce, environmental studies, financial markets, multimedia data processing, network monitoring, and social service analysis.
    \item \textsc{Fundamentals Artificial Intelligence} provides an introduction to the core problem-solving and knowledge representation paradigms in artificial intelligence. It covers Lisp programming with regular assignments, as well as topics such as search methods, planning techniques, knowledge structures, natural language processing, expert systems, vision, and parallel architectures.
    \item \textsc{Differential Equations} covers various topics in differential equations, including first-order and second-order linear equations with constant coefficients, power series solutions, and linear systems. Students will explore the principles and applications of these mathematical concepts.
\end{itemize}

A detailed statistics of the exam dataset is summarized in \cref{tab:new-exam}.
The experiment results of exam dataset are provided in \cref{exam_result}.
\input{tables/new_exam}
\input{tables/exam_result}

\section{Experimental Details}
\label{appendix:details}

\subsection{Prompts}
\label{prompt}
The APIs of ChatGPT and GPT-4 have three message parameters: \verb|SYSTEM|, \verb|USER|, and \verb|ASSISTANT|. The \verb|SYSTEM| parameter represents the system prompt, which provides context and instructions to the model. The \verb|USER| parameter is the training prompt or input provided by the user, and the \verb|ASSISTANT| parameter contains the output of the model or the response. All system prompts and training prompts used in our experiments are provided below.

\begin{mdframed}
\textbf{System Prompt for Zero-Shot, Few-Shot, and Chain-of-Thought settings.}

Please provide a clear and step-by-step solution for a scientific problem in the categories of Chemistry, Physics, or Mathematics. The problem will specify the unit of measurement, which should not be included in the answer. Express the final answer as a decimal number with three digits after the decimal point. Conclude the answer by stating "The answer is therefore \verb|\boxed{[ANSWER]}|."
\end{mdframed}

\begin{mdframed}
\textbf{System Prompt for Few-Shot Learning + Python.}

Please provide a clear and step-by-step solution for a scientific problem in the categories of Chemistry, Physics, or Mathematics. The problem will specify the unit of measurement. Please translate the solution steps into Python code and encase the Python code within triple backticks for clarity.
\end{mdframed}

\begin{mdframed}
\textbf{System Prompt for Few-Show Learning + Wolfram Language.}

Please provide a clear and step-by-step solution for a scientific problem in the categories of Chemistry, Physics, or Mathematics. The problem will specify the unit of measurement. Please translate the solution steps into Wolfram code and encase the Wolfram Language code within triple backticks for clarity.
\end{mdframed}

\begin{mdframed}
\textbf{System Prompt for Evaluation Protocol.}

Examine the given problem, the correct solution, and the model's solution. Identify the reason for the error in the model's solution based on the following 10 categories:

1. Logical Decomposition and Analysis Skills: This ability involves decomposing the problem into smaller, manageable parts, and understanding the relationships between these parts.

2. Identification of Assumptions: This skill involves the AI's ability to recognize relevant and necessary assumptions in the problem.

3. Spatial Perception: This is important for understanding problems in areas such as physics and chemistry, where you need to visualize molecules, forces, fields, etc.

4. Causal Reasoning: This is the ability to understand cause and effect relationships.

5. Problem Deduction Skills: This pertains to the ability to infer and deduce potential solutions or underlying principles from the given information in a problem.

6. Abstract Reasoning: This skill involves the ability to understand complex concepts that can't be perceived physically, and to recognize patterns or relationships beyond concrete examples.

7. Scientific Literacy:  This skill involves a comprehensive understanding of key scientific principles, terminology, and methodologies across a range of disciplines.

8. Code Conversion Skills: This denotes the ability to accurately translate solution steps into different programming languages, like Python or Wolfram, without syntax errors.

9. Logical Reasoning: This is the ability to make a reasoned argument and to identify fallacies or inconsistencies in an argument or set of data.

10. Calculation Skills: This involves the ability to accurately carry out mathematical operations and computations.

Conclude your final error reason category number within \verb|\boxed{}|.
\end{mdframed}

\begin{mdframed}
\textbf{Training Prompt for Zero-Shot Chain-of-Thought.}

\textit{Stage 1:}

\textcolor{blue}{Input:} [Input-Question] Let's think step by step.

\textcolor{blue}{Output:} \verb|<explanation>|

\textit{Stage 2:}

\textcolor{blue}{Input:} [Input-Question] Let's think step by step. [Explanation]. Therefore, the answer is:

\textcolor{blue}{Output:} \verb|<answer>|
\end{mdframed}

\begin{mdframed}
\textbf{Training Prompt for Few-Shot Chain-of-Thought.}

\textcolor{blue}{Input:}

Problem 1: [Question 1] Explanation for Problem 1: [Explanation 1]. The answer is \verb|\boxed{[Answer 1]}|.

Problem 2: [Question 2] Explanation for Problem 2: [Explanation 2]. The answer is \verb|\boxed{[Answer 2]}|.

...

Problem n: [Question n] Explanation for Problem n: [Explanation n]. The answer is \verb|\boxed{[Answer n]}|.

Problem n+1: [Question n+1]

\textcolor{blue}{Output:} Explanation for Problem n+1: \verb|<explanation>|. The answer is \verb|\boxed{<answer>}|.
\end{mdframed}

\begin{mdframed}
\textbf{Training Prompt for Few-Shot Python or Wolfram Language.}

\textcolor{blue}{Input:}

Problem 1: [Question 1] Explanation for Problem 1: [Explanation 1]. Python/Wolfram language for Problem 1: \verb|```[Python/Wolfram code 1]```|.

Problem 2: [Question 2] Explanation for Problem 2: [Explanation 2]. Python/Wolfram language for Problem 2: \verb|```[Python/Wolfram code 2]```|.

...

Problem n: [Question n] Explanation for Problem n: [Explanation n]. Python/Wolfram language for Problem n: \verb|```[Python/Wolfram code n]```|.

Problem n+1: [Question n+1]

\textcolor{blue}{Output:} Explanation for Problem n+1: \verb|<explanation>|. Python/Wolfram language for Problem n+1: \verb|```[Python/Wolfram code n+1]```|.
\end{mdframed}

\begin{mdframed}
\textbf{Training Prompt for Evaluation Protocol.}

\textcolor{blue}{Input:} The question is [input-question]. The correct solution is [Correct-Solution]. The model solution is [Model-Solution].

\textcolor{blue}{Output:} \verb|<Error Type>|
\end{mdframed}

\begin{mdframed}
\textbf{Training Prompt for Evaluation Protocol in Python or Wolfram Language.}

\textcolor{blue}{Input:} The question is [input-question]. The correct solution is [Correct-Solution]. The model solution is [Model-Solution]. The translated program generates the answer as [Program Generated Answer], which is treated as model's output answer.

\textcolor{blue}{Output:} \verb|<Error Type>|
\end{mdframed}

\subsection{Implementation Details}
All model output is extracted using $\backslash$boxed\{\} notation. To prevent any missed extractions, we supplement this process with a manual check. For both Python and Wolfram settings, we extract the programming language with the triple backtick \verb|```|, subsequently executing it within the corresponding language. The entirety of our code can be accessed via \href{https://github.com/mandyyyyii/scibench}{this repository}.

\begin{figure}[b]
	\centering
	\includegraphics[width=0.65\linewidth]{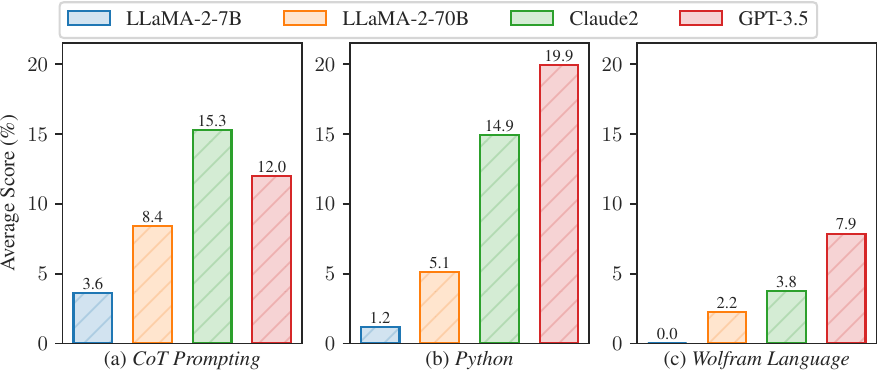}
	\caption{Comparison between few-shot learning with external tools.}
	\label{fig:wolfram}
\end{figure}
\begin{figure}[b]
	\centering
	\includegraphics[width=0.9\linewidth]{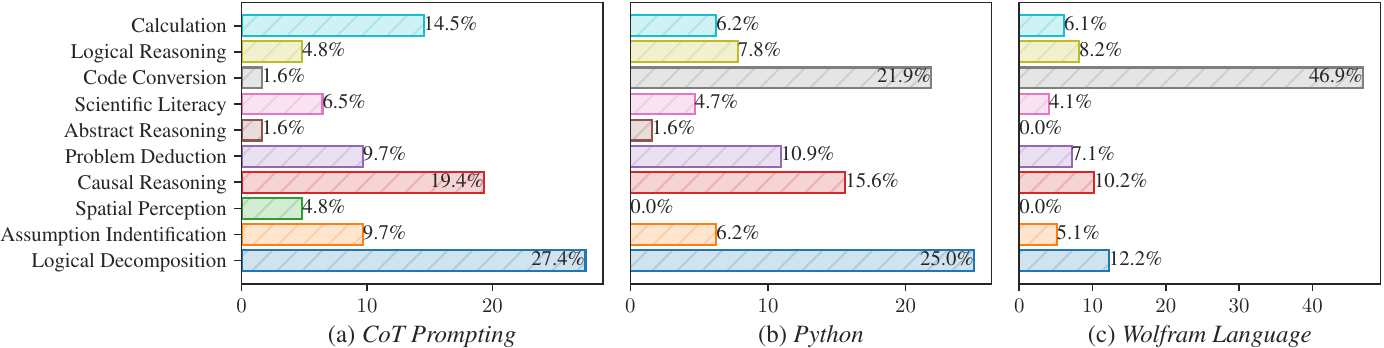}
	\caption{Error Analysis between few-shot learning with external tools.}
	\label{fig:wolfram-analysis}
\end{figure}

\subsection{Additional Experiment on Wolfram Language}
\label{appendix:wolfram}

The experiment results and error analysis for using Wolfram Language as external tools are presented in Figure \ref{fig:wolfram} and Figure \ref{fig:wolfram-analysis}, compared with using CoT and Python Language. We observe that the use of external tools can weaken other skills, particularly the code conversion skills. This issue becomes particularly prominent when using the Wolfram Language, with 46.9\% error rate in code conversion skill. Despite providing grammar specifications in system prompts and a few examples as demonstrations, most attempts of code conversion result in syntax errors. In Wolfram Language, the error mainly comes from the violation of variable rules (for instance, Wolfram Language reserves certain letters such as $E$ as protected symbols and disallows underscores in variable names) or incorrect usage of certain functions. This observation suggests a potential improvement for LLM when using Wolfram Language. 

\section{Problem Solving Abilities of Current LLMs}
\label{reasoning}
\subsection{Assessment of the Evaluation Protocol}
In order to assess the effectiveness of our evaluation protocol's classification, we enlisted the assistance of two annotators to determine whether the errors identified by the model verifier were accurate or not. Through the annotation of 151 samples across different settings, we observed that 123 of them were correctly classified, resulting in an accuracy rate of 81.45$\%$. Two human annotators participate in the process. Decisions on the final abilities are determined by annotators, aided by assistants. By going through errors, these two annotators develop ten abilities and then employ a Language Learning Model (LLM) as a third evaluator to suggest additional abilities. They then compare and refine their findings based on this input. Ultimately, the final outcomes are determined by the annotators. After LLM annotate the error reasons, we conduct human-check by sampling 151 examples across all settings to make sure the annotations make sense. We make this human-AI cooperated analysis pipeline to reduce the cost of human post-analysis, while incorporate human checking to make sure the correctness of LLM decision and try to reduce the risk that reviewer mentioned. Though not perfect, we believe it can be another type of analyzing framework for future study of LLM problem-solving.

\subsection{Examples}

\input{tables/appendix_eg}

In the context of each specific capability, we present several exemplary errors accompanied by their corresponding classifications and explanations derived from the GPT model. Referencing Figure \ref{appendix:eg_error}, the ChatGPT solution employing the Chain-of-Thought method corresponds to error category 4, "Causal Reasoning". The model explanation posits that "The error reason category is 4. Causal Reasoning. The model solution uses the relativistic momentum formula to calculate the momentum of the electron, which is not necessary for this problem since the electron is traveling at only $1.00 \%$ of the speed of light. The relativistic momentum formula is only needed when the velocity of the particle is a significant fraction of the speed of light. Therefore, using this formula in this problem leads to an incorrect value for the momentum and consequently an incorrect value for the de Broglie wavelength." Meanwhile, the ChatGPT solution that doesn't utilize the Chain-of-Thought method is categorized under error case 10, "Calculation Skills". This example is further highlighted as an illustrative showcase for explanations of the Chain-of-Thought method in Section \ref{sec_ana}. More examples resulting in different error types are shown in \cref{appendix:error_cap8,appendix:error_cap3,appendix:error_cap4,appendix:error_cap1,appendix:error_cap6,appendix:error_cap7,appendix:error_cap5,appendix:error_cap10,appendix:error_cap9,appendix:error_cap11}. 

\input{tables/error_cap8}
\input{tables/error_cap3}
\input{tables/error_cap4}
\input{tables/error_cap1}
\input{tables/error_cap6}
\input{tables/error_cap7}
\input{tables/error_cap5}
\input{tables/error_cap10}
\input{tables/error_cap9}
\input{tables/error_cap11}

%% file: tables/subj_eg.tex
\begin{figure}[h!]
\fbox{\begin{minipage}{0.98\textwidth}
\scriptsize
\textbf{Problem} (\textcolor{brown}{\texttt{fund}})\\
Two charged particles are fixed to an $x$ axis: Particle 1 of charge $q_1=2.1 \times 10^{-8} \mathrm{C}$ is at position $x=20 \mathrm{~cm}$ and particle 2 of charge $q_2=-4.00 q_1$ is at position $x=70 \mathrm{~cm}$. At what coordinate on the axis (other than at infinity) is the net electric field produced by the two particles equal to zero?
 \\
\textbf{Answer:} $-30 \mathrm{~cm}$

\end{minipage}}
\fbox{\begin{minipage}{0.98\textwidth}
\scriptsize
\textbf{Problem} (\textcolor{brown}{\texttt{thermo}})\\
$\mathrm{N}_2 \mathrm{O}_3$ dissociates according to the equilibrium $\mathrm{N}_2 \mathrm{O}_3(\mathrm{~g}) \rightleftharpoons \mathrm{NO}_2(\mathrm{~g})+\mathrm{NO}(\mathrm{g})$. At $298 \mathrm{~K}$ and one bar pressure, the degree of dissociation defined as the ratio of moles of $\mathrm{NO}_2(g)$ or $\mathrm{NO}(g)$ to the moles of the reactant assuming no dissociation occurs is $3.5 \times 10^{-3}$. Calculate $\Delta G_R^{\circ}$ for this reaction.\\
\textbf{Answer:} 28 $\mathrm{~kJ} \mathrm{~mol}^{-1}$ 

\end{minipage}}
\fbox{\begin{minipage}{0.98\textwidth}
\scriptsize
\textbf{Problem} (\textcolor{brown}{\texttt{class}})\\
Halley's comet, which passed around the sun early in 1986, moves in a highly elliptical orbit with an eccentricity of 0.967 and a period of 76 years. Calculate its minimum distances from the Sun.\\
\textbf{Answer:} 8.8 $\times 10^{10} \mathrm{m}$
\end{minipage}}
\fbox{\begin{minipage}{0.98\textwidth}
\scriptsize
\textbf{Problem} (\textcolor{brown}{\texttt{quan}})\\
A one-particle, one-dimensional system has $\Psi=a^{-1 / 2} e^{-|x| / a}$ at $t=0$, where $a=1.0000 \mathrm{~nm}$. At $t=0$, the particle's position is measured. Find the probability that the measured value is between $x=0$ and $x=2 \mathrm{~nm}$.

\textbf{Answer:} 0.4908

\end{minipage}}
\fbox{\begin{minipage}{0.98\textwidth}
\scriptsize
\textbf{Problem} (\textcolor{brown}{\texttt{chemmc}})\\
One of the most powerful modern techniques for studying structure is neutron diffraction. This technique involves generating a collimated beam of neutrons at a particular temperature from a high-energy neutron source and is accomplished at several accelerator facilities around the world. If the speed of a neutron is given by $v_{\mathrm{n}}=\left(3 k_{\mathrm{B}} T / m\right)^{1 / 2}$, where $m$ is the mass of a neutron, then what temperature is needed so that the neutrons have a de Broglie wavelength of $50 \mathrm{pm}$ ?

\textbf{Answer:} 2500 $\mathrm{K}$ 

\end{minipage}}
\fbox{\begin{minipage}{0.98\textwidth}
\scriptsize
\textbf{Problem} (\textcolor{brown}{\texttt{atkins}})\\
The change in molar internal energy when $\mathrm{CaCO}_3(\mathrm{~s})$ as calcite converts to another form, aragonite, is $+0.21 \mathrm{~kJ} \mathrm{~mol}^{-1}$. Calculate the difference between the molar enthalpy and internal energy changes when the pressure is 1.0 bar given that the densities of the polymorphs are $2.71 \mathrm{~g} \mathrm{~cm}^{-3}$ and $2.93 \mathrm{~g} \mathrm{~cm}^{-3}$, respectively.\\
\textbf{Answer: }-0.28$\mathrm{~Pa} \mathrm{~m}^3 \mathrm{~mol}^{-1}$
\end{minipage}}
\fbox{\begin{minipage}{0.98\textwidth}
\scriptsize
\textbf{Problem} (\textcolor{brown}{\texttt{matter}})\\
In an industrial process, nitrogen is heated to $500 \mathrm{~K}$ at a constant volume of $1.000 \mathrm{~m}^3$. The gas enters the container at $300 \mathrm{~K}$ and $100 \mathrm{~atm}$. The mass of the gas is $92.4 \mathrm{~kg}$. Use the van der Waals equation to determine the approximate pressure of the gas at its working temperature of $500 \mathrm{~K}$. For nitrogen, $a=1.39 \mathrm{dm}^6 \mathrm{~atm} \mathrm{~mol}^{-2}, b=0.0391 \mathrm{dm}^3 \mathrm{~mol}^{-1}$.\\
\textbf{Answer:} 140 $\mathrm{~atm}$
\end{minipage}}
\fbox{\begin{minipage}{0.98\textwidth}
\scriptsize
\textbf{Problem} (\textcolor{brown}{\texttt{calc}})\\
A planning engineer for a new alum plant must present some estimates to his company regarding the capacity of a silo designed to contain bauxite ore until it is processed into alum. The ore resembles pink talcum powder and is poured from a conveyor at the top of the silo. The silo is a cylinder $100 \mathrm{ft}$ high with a radius of $200 \mathrm{ft}$. The conveyor carries ore at a rate of $60,000 \pi \mathrm{~ft}^3 / \mathrm{h}$ and the ore maintains a conical shape whose radius is 1.5 times its height. If, at a certain time $t$, the pile is $60 \mathrm{ft}$ high, how long will it take for the pile to reach the top of the silo?\\
\textbf{Answer:} 9.8 $\mathrm{h}$
\end{minipage}}
\fbox{\begin{minipage}{0.98\textwidth}
\scriptsize
\textbf{Problem} (\textcolor{brown}{\texttt{stat}})\\
In a study concerning a new treatment of a certain disease, two groups of 25 participants in each were followed for five years. Those in one group took the old treatment and those in the other took the new treatment. The theoretical dropout rate for an individual was $50 \%$ in both groups over that 5 -year period. Let $X$ be the number that dropped out in the first group and $Y$ the number in the second group. Assuming independence where needed, give the sum that equals the probability that $Y \geq X+2$. HINT: What is the distribution of $Y-X+25$ ?

\textbf{Answer:} 0.3359

\end{minipage}}
\fbox{\begin{minipage}{0.98\textwidth}
\scriptsize
\textbf{Problem} (\textcolor{brown}{\texttt{diff}})\\
Newton's law of cooling states that the temperature of an object changes at a rate proportional to the difference between its temperature and that of its surroundings. Suppose that the temperature of a cup of coffee obeys Newton's law of cooling. If the coffee has a temperature of $200^{\circ} \mathrm{F}$ when freshly poured, and $1 \mathrm{~min}$ later has cooled to $190^{\circ} \mathrm{F}$ in a room at $70^{\circ} \mathrm{F}$, determine when the coffee reaches a temperature of $150^{\circ} \mathrm{F}$\\
\textbf{Answer:} 6.07 min

\end{minipage}}
\vspace{-1mm}
\caption{ Textbook examples with acronym highlighted in \textcolor{brown}{brown}.}
\label{appendix:subj}
\end{figure}

%% file: tables/img_eg0.tex
\begin{figure}[h]
\fbox{\begin{minipage}{0.98\textwidth}
\scriptsize
\textbf{Problem}\\
The region $\mathscr{R}$ enclosed by the curves $y=x$ and $y=x^2$ is rotated about the $x$-axis. Find the volume of the resulting solid. \\
\end{minipage}}
\fbox{\begin{minipage}{0.98\textwidth}
\scriptsize
\textbf{Image}\\
\includegraphics[width=0.5\textwidth]{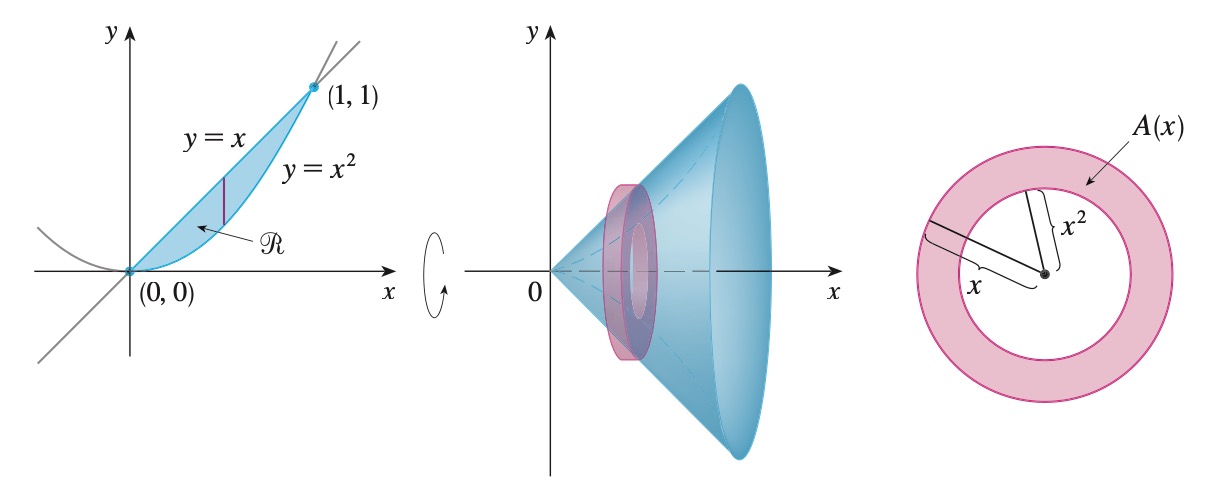}
\end{minipage}}
\fbox{\begin{minipage}{0.98\textwidth}
\scriptsize
\textbf{Correct Solution}\\
The curves $y=x$ and $y=x^2$ intersect at the points $(0,0)$ and $(1,1)$. The region between them, the solid of rotation, and a cross-section perpendicular to the $x$-axis are shown in the Figure. A cross-section in the plane $P_x$ has the shape of a washer (an annular ring) with inner radius $x^2$ and outer radius $x$, so we find the cross-sectional area by subtracting the area of the inner circle from the area of the outer circle:$$A(x)=\pi x^2-\pi\left(x^2\right)^2=\pi\left(x^2-x^4\right)$$ Therefore we have$$\begin{aligned} V & =\int_0^1 A(x) d x=\int_0^1 \pi\left(x^2-x^4\right) d x \\& =\pi\left[\frac{x^3}{3}-\frac{x^5}{5}\right]_0^1=\frac{2 \pi}{15}\end{aligned}$$\\
\textbf{Final Answer:} $ \frac{2\pi}{15} $
\end{minipage}}
\vspace{-1mm}
\caption{The example from the textbook \textit{Calculus: Early Transcendentals}.}
\label{img_eg0}
\end{figure}

%% file: tables/img_eg1.tex
\begin{figure}
\fbox{\begin{minipage}{0.98\textwidth}
\scriptsize
\textbf{Problem}\\
A $2.00 \mathrm{~kg}$ particle moves along an $x$ axis in one-dimensional motion while a conservative force along that axis acts on it. The potential energy $U(x)$ associated with the force is plotted in the Figure. That is, if the particle were placed at any position between $x=0$ and $x=7.00 \mathrm{~m}$, it would have the plotted value of $U$. At $x=6.5 \mathrm{~m}$, the particle has velocity $\vec{v}_0=(-4.00 \mathrm{~m} / \mathrm{s}) \hat{\mathrm{i}}$. From the Figure, determine the particle's speed at $x_1=4.5 \mathrm{~m}$.
\end{minipage}}

\fbox{\begin{minipage}{0.98\textwidth}
\scriptsize
\textbf{Image}\\
\includegraphics[width=0.2\textwidth]{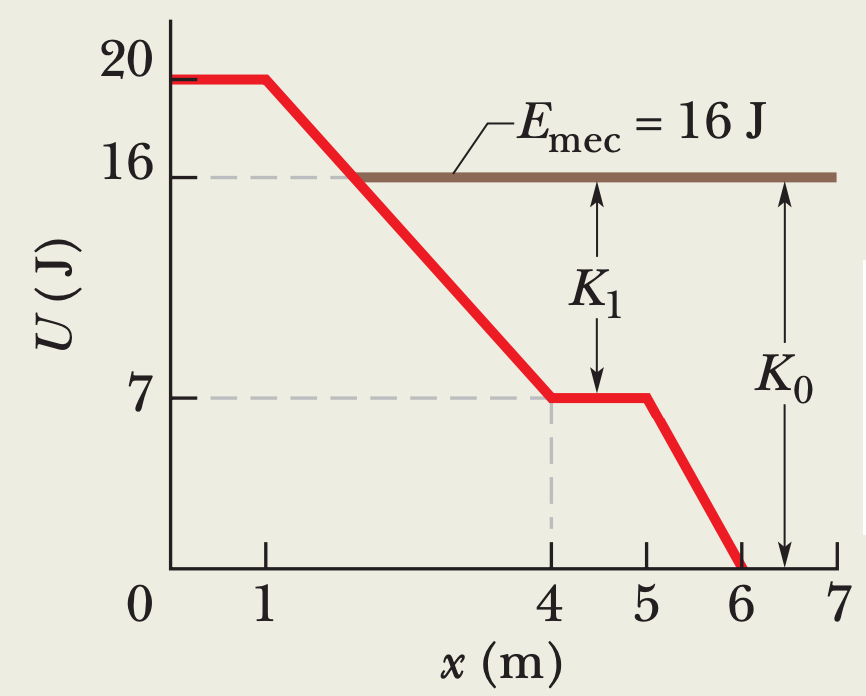}
\end{minipage}}

\fbox{\begin{minipage}{0.98\textwidth}
\scriptsize
\textbf{Correct Solution}\\
The particle's kinetic energy is given by Eq $\left(K=\frac{1}{2} m v^2\right)$. Because only a conservative force acts on the particle, the mechanical energy $E_{\mathrm{mec}}(=K+U)$ is conserved as the particle moves. Therefore, on a plot of $U(x)$, the kinetic energy is equal to the difference between $E_{\mathrm{mec}}$ and $U$.\\
Calculations: At $x=6.5 \mathrm{~m}$, the particle has kinetic energy
\begin{align}
	K_0 & =\frac{1}{2} m v_0^2=\frac{1}{2}(2.00 \mathrm{~kg})(4.00 \mathrm{~m} / \mathrm{s})^2 \\
	& =16.0 \mathrm{~J} .
\end{align}
Because the potential energy there is $U=0$, the mechanical energy is $E_{\mathrm{mec }}=K_0+U_0=16.0 \mathrm{~J}+0=16.0 \mathrm{~J}$. This value for $E_{\mathrm{mec}}$ is plotted as a horizontal line in the Figure. From that figure we see that at $x=4.5 \mathrm{~m}$, the potential energy is $U_1=7.0 \mathrm{~J}$. The kinetic energy $K_1$ is the difference between $E_{\mathrm{mec}}$ and $U_1$ :
\begin{align}
	K_1=E_{\mathrm{mec}}-U_1=16.0 \mathrm{~J}-7.0 \mathrm{~J}=9.0 \mathrm{~J}.	
\end{align}
Because $K_1=\frac{1}{2} m v_1^2$, we find $v_1=3.0 \mathrm{~m} / \mathrm{s}$.\\
\textbf{Final Answer:} $3.0 \mathrm{~m} / \mathrm{s}$
\end{minipage}}
\vspace{-1mm}
\caption{An example problem from the textbook \textit{Fundamentals of Physics}.}
\label{image_eg1}
\end{figure}

%% file: tables/img_eg2.tex
\begin{figure}
\fbox{\begin{minipage}{0.98\textwidth}	
\scriptsize
\textbf{Problem}\\
If the particles in a system all move together, the com moves with them-no trouble there. But what happens when they move in different directions with different accelerations? Here is an example. The three particles in the Figure are initially at rest. Each experiences an external force due to bodies outside the three-particle system. The directions are indicated, and the magnitudes are $F_1=6.0 \text{~N}, F_2=12 \text{~N}$, and $F_3=14 \text{~N}$. What is the acceleration of the center of mass of the system? \\
\end{minipage}}

\fbox{\begin{minipage}{0.98\textwidth}
\scriptsize
\textbf{Image}\\
\includegraphics[width=.3\textwidth]{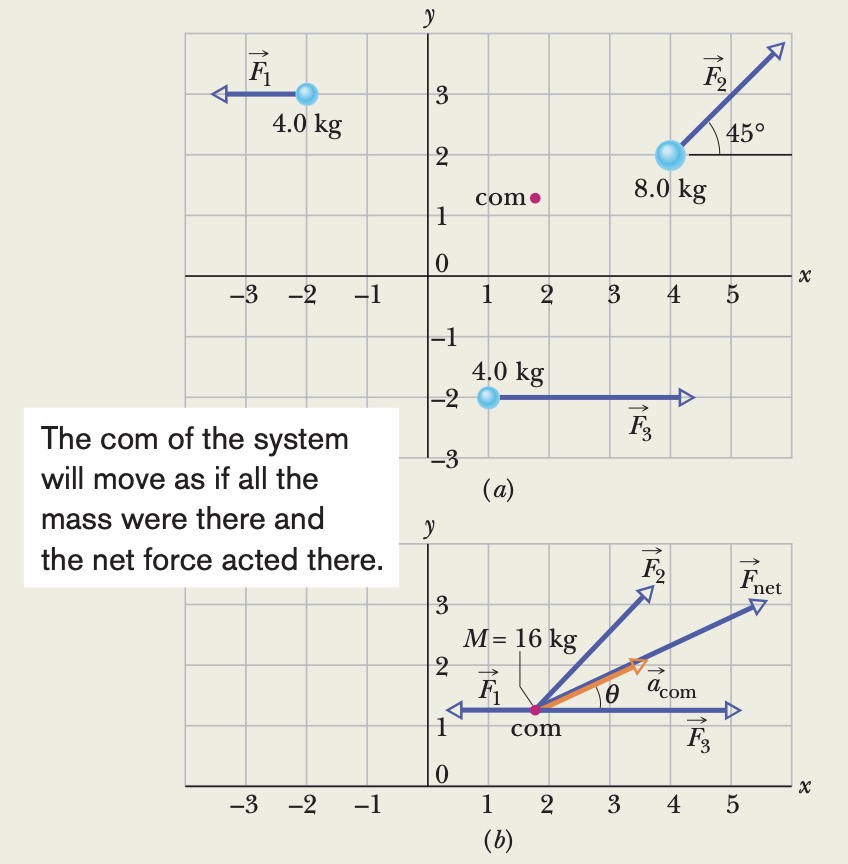}
\end{minipage}}

\fbox{\begin{minipage}{0.98\textwidth}
\scriptsize
\textbf{Correct Solution}\\
The position of the center of mass is marked by a dot in the figure. We can treat the center of mass as if it were a real particle, with a mass equal to the system's total mass $M=16 \mathrm{~kg}$. We can also treat the three external forces as if they act at the center of mass (Figure b). We can now apply Newton's second law $\left(\vec{F}_{\mathrm{net }}=m \vec{a}\right)$ to the center of mass, writing
\begin{align}
	\vec{F}_{\mathrm{net}} &=M \vec{a}_{\mathrm{com}},\\
	\vec{F}_1+\vec{F}_2+\vec{F}_3 &=M \vec{a}_{\mathrm{com}}, \\ \vec{a}_{\mathrm{com}}=\frac{\vec{F}_1+\vec{F}_2+\vec{F}_3}{M}.
\end{align}
The equation tells us that the acceleration $\vec{a}_{\mathrm{com}}$ of the center of mass is in the same direction as the net external force $\vec{F}_{\mathrm {net }}$ on the system (Figure b). Because the particles are initially at rest, the center of mass must also be at rest. As the center of mass then begins to accelerate, it must move off in the common direction of $\vec{a}_{\mathrm{com}}$ and $\vec{F}_{\mathrm{net}}$. We can evaluate the right side of Eq. S6 directly on a vector-capable calculator, or we can rewrite Eq. S6 in component form, find the components of $\vec{a}_{\mathrm{com}}$, and then find $\vec{a}_{\text {com }}$. Along the $x$ axis, we have
\begin{align}
	a_{\mathrm{com}, x} =\frac{F_{1 x}+F_{2 x}+F_{3 x}}{M} =\frac{-6.0 \mathrm{~N}+(12 \mathrm{~N}) \cos 45^{\circ}+14 \mathrm{~N}}{16 \mathrm{~kg}}=1.03 \mathrm{~m} / \mathrm{s}^2.
\end{align}
Along the $y$ axis, we have
\begin{align}
	a_{\mathrm{com}, y} =\frac{F_{1 y}+F_{2 y}+F_{3 y}}{M} =\frac{0+(12 \mathrm{~N}) \sin 45^{\circ}+0}{16 \mathrm{~kg}}=0.530 \mathrm{~m} / \mathrm{s}^2.
\end{align}
From these components, we find that $\vec{a}_{\mathrm{com}}$ has the magnitude
\begin{align}
	a_{\mathrm{com}} =\sqrt{\left(a_{\mathrm{com}, x}\right)^2+\left(a_{\mathrm{com}, y}\right)^2} =1.16 \mathrm{~m} / \mathrm{s}^2.
\end{align}\\
\textbf{Final Answer:} $1.16 \mathrm{~m} / \mathrm{s}^2$
\end{minipage}}
\vspace{-1mm}
\caption{The example from the textbook \textit{Fundamentals of Physics}.}
\label{img_eg2}
\end{figure}

%% file: tables/img_eg3.tex
\begin{figure}
\fbox{\begin{minipage}{0.98\textwidth}
\scriptsize
\textbf{Problem}\\
At time $t=0$ a tank contains $Q_0 \mathrm{lb}$ of salt dissolved in 100 gal of water; see Figure 2.3.1. Assume that water containing $\frac{1}{4} \mathrm{lb}$ of salt/gal is entering the tank at a rate of $r \mathrm{gal} / \mathrm{min}$ and that the well-stirred mixture is draining from the tank at the same rate. Set up the initial value problem that describes this flow process. By finding the amount of salt $Q(t)$ in the tank at any time, and the limiting amount $Q_L$ that is present after a very long time, if $r=3$ and $Q_0=2 Q_L$, find the time $T$ after which the salt level is within $2 \%$ of $Q_L$. \\
\end{minipage}}

\fbox{\begin{minipage}{0.98\textwidth}
\scriptsize
\textbf{Image}\\
\includegraphics[width=0.3\textwidth]{figures/img_question3.png}
\end{minipage}}

\fbox{\begin{minipage}{0.98\textwidth}
\scriptsize
\textbf{Correct Solution}\\
We assume that salt is neither created nor destroyed in the tank. Therefore variations in the amount of salt are due solely to the flows in and out of the tank. More precisely, the rate of change of salt in the tank, $d Q / d t$, is equal to the rate at which salt is flowing in minus the rate at which it is flowing out. In symbols,$$\frac{d Q}{d t}=\text{ rate in }- \text{ rate out }$$The rate at which salt enters the tank is the concentration $\frac{1}{4} \mathrm{lb} / \mathrm{gal}$ times the flow rate $r \mathrm{gal} / \mathrm{min}$, or $(r / 4) \mathrm{lb} / \mathrm{min}$. To find the rate at which salt leaves the tankl we need to multiply the concentration of salt in the tank by the rate of outflow, $r \mathrm{gal} / \mathrm{min}$. Since the rates of flow in and out are equal, the volume of water in the tank remains constant at $100 \mathrm{gal}$, and since the mixture is "well-stirred," the concentration throughout the tank is the same, namely, $[Q(t) / 100] \mathrm{lb} / \mathrm{gal}$.Therefore the rate at which salt leaves the tank is $[r Q(t) / 100] \mathrm{lb} / \mathrm{min}$. Thus the differential equation governing this process is$$\frac{d Q}{d t}=\frac{r}{4}-\frac{r Q}{100}$$The initial condition is$$Q(0)=Q_0$$Upon thinking about the problem physically, we might anticipate that eventually the mixture originally in the tank will be essentially replaced by the mixture flowing in, whose concentration is $\frac{1}{4} \mathrm{lb} / \mathrm{gal}$. Consequently, we might expect that ultimately the amount of salt in the tank would be very close to $25 \mathrm{lb}$. We can also find the limiting amount $Q_L=25$ by setting $d Q / d t$ equal to zero and solving the resulting algebraic equation for $Q$. Rewriting it in the standard form for a linear equation, we have$$\frac{d Q}{d t}+\frac{r Q}{100}=\frac{r}{4}$$Thus the integrating factor is $e^{r t / 100}$ and the general solution is$$Q(t)=25+c e^{-r t / 100}$$where $c$ is an arbitrary constant. To satisfy the initial condition, we must choose $c=Q_0-25$. Therefore the solution of the initial value problem is $$Q(t)=25+(Q_0-25) e^{-r t / 100}$$ $$Q(t)=25(1-e^{-r t / 100})+Q_0 e^{-r t / 100}$$From above Equations, you can see that $Q(t) \rightarrow 25$ (lb) as $t \rightarrow \infty$, so the limiting value $Q_L$ is 25 , confirming our physical intuition. Further, $Q(t)$ approaches the limit more rapidly as $r$ increases. In interpreting the solution, note that the second term on the right side is the portion of the original salt that remains at time $t$, while the first term gives the amount of salt in the tank due to the action of the flow processes. Now suppose that $r=3$ and $Q_0=2 Q_L=50$; then$$Q(t)=25+25 e^{-0.03 t}$$Since $2 \%$ of 25 is 0.5 , we wish to find the time $T$ at which $Q(t)$ has the value 25.5. Substituting $t=T$ and $Q=25.5$ and solving for $T$, we obtain$$T=(\ln 50) / 0.03 \cong 130.400766848(\mathrm{~min}).$$
\textbf{Final Answer:} $(\ln 50) / 0.03$

\end{minipage}}
\caption{The example from the textbook \textit{Elementary Differential Equations and Boundary
    Value Problems}.}
\label{img_eg3}
\end{figure}

%% file: tables/new_exam.tex
\begin{table*}
\caption{Statistics of the close exam dataset. We report the number of problem instances in each exam and the ratio of problems in the exam that include detailed solutions. We further report the ratio of problems in different formats, including free-response, multiple-choice, and true-false. For reference, the number in parentheses denotes the grading points assigned to the problems.}
\centering
\small
\label{tab:new-exam}
	\SetTblrInner{rowsep=1pt}
	\begin{tblr}{
	colspec = {lccccccc},
	row{1-2} = {bg=gray!25},
	row{even[3-7]} = {bg=gray!5}
	}
\toprule
  & \SetCell[c=2]{c}{\textit{Data Mining}} & & \SetCell[c=2]{c}{\textit{Machine Learning}} & & \SetCell[c=3]{c}{\textit{Differential Equations}}       \\
\cmidrule[lr]{2-3}\cmidrule[lr]{4-5}\cmidrule[lr]{6-8}
                       & Midterm & Final & Midterm & Final & Exam 1  & Exam 2  & Final  \\\midrule
\# Problems                    &  25 (90)     & 24 (75) &    12 (56)   & 16 (75) &   8 (100) & 8 (100) & 11 (95) \\
\% Solutions             &  56.0\% (58)      &    16.7\% (19)    &  100.0\% (56)   &   31.2\% (26)   &   100.0\% (100)    & 100.0\% (100)      & 90.9\% (90)   \\
\% Free-response &  40.0\% (46)      &    33.3\% (29)    &  66.7\% (38)   &   81.3\% (62)   &   100.0\% (100)    & 100.0\% (100)      & 90.9\% (90)   \\
\% Multiple-choice &  28.0\% (28)      &    29.2\% (28)    &  33.3\% (18)   &   18.7\% (13)   &   0.0\% (0)    & 0.0\% (0)      & 9.1\% (5)   \\
\% True-false &  32.0\% (16)      &    37.5\% (18)    &  0.0\% (0)   &   0.0\% (0)   &   0.0\% (0)    & 0.0\% (0)      & 0.0\% (0)   \\
\bottomrule
\end{tblr}
\vskip-1em
\end{table*}

%% file: tables/exam_result.tex
\begin{table}
\caption{Experimental results in terms of total scores under zero-shot learning on the exam dataset. The best performing score is highlighted in \textbf{bold} and second-best is \ul{underlined}.}
\label{exam_result}
\centering
\resizebox{0.7\textwidth}{!}{
\SetTblrInner{rowsep=0.8pt}
	\begin{tblr}{
	colspec = {ccccccccc},
	row{1-2} = {bg=gray!25},
	row{5-6,9-10} = {bg=gray!10},
	}
\toprule
\SetCell[r=2]{c}{Model} & \SetCell[r=2]{c}{Setting}  & \SetCell[c=2]{c}{\textit{Data Mining}} & & \SetCell[c=2]{c}{\textit{Machine Learning}} & & \SetCell[c=3]{c}{\textit{Differential Equations}}       \\\cmidrule[lr]{3-4}\cmidrule[lr]{5-6}\cmidrule[lr]{7-9}
                      & & Midterm & Final & Midterm & Final & Exam 1  & Exam 2  & Final  \\\midrule
\SetCell[r=2]{c}LLaMA-2-7B & Zero    &     24 / 90    &     14 / 75    &    6 / 56   &      6/ 75   &      5 / 100    &  0 / 100      &  0 / 95   \\
 & Zero$+$CoT    &     18 / 90    &     14 / 75    &    2 / 56   &      10 / 75   &      10 / 100    &  0 / 100      &  10 / 95   \\
\SetCell[r=2]{c}LLaMA-2-70B & Zero    &     23 / 90    &     18 / 75    &    18 / 56   &      12 / 75   &      20 / 100    &  5 / 100      &  0 / 95   \\
 & Zero$+$CoT    &     31 / 90    &     18 / 75    &   10 / 56   &      11/ 75   &      \ul{35} / 100    &  10 / 100      &  0 / 95   \\
\SetCell[r=2]{c}Claude2 & Zero    &     37 / 90    &     26 / 75    &    28 / 56   &  35 / 75   &      \ul{35} / 100    &  30 / 100      &  \ul{20} / 95   \\
 & Zero$+$CoT    &     33 / 90    &     38 / 75    &    22 / 56   &      \textbf{41} / 75   &      25 / 100    &  15 / 100      &  \ul{20} / 95   \\
\SetCell[r=2]{c}GPT-3.5 & Zero    &    44 / 90    &    \ul{39} / 75    &   16 / 56   &     32 / 75   &     0 / 100    & \ul{45} / 100      & 15 / 95   \\
 & Zero$+$CoT    &   38 / 90     &   33 / 75   &    \ul{32} / 56 &    \ul{37} / 75   &    28 / 100   &  30 / 100    &   10 / 95\\ 
\SetCell[r=2]{c}GPT-4 & Zero  &   \ul{56} / 90     &   \textbf{44} / 75   &    30 / 56 &    \ul{37} / 75    &    25 / 100   &  \textbf{80} / 100    &   \textbf{25} / 95\\ 
 & Zero$+$CoT           &   \textbf{58} / 90     &   32 / 75   &    \textbf{40} / 56 &    35 / 75    &    \textbf{50} / 100   &  70 / 100    &   15 / 95\\ \bottomrule
\end{tblr}
}
\end{table}

%% file: tables/appendix_eg.tex
\begin{figure}[h]

\fbox{\begin{minipage}{0.98\textwidth}
\scriptsize
\textbf{Problem}\\
Calculate the de Broglie wavelength of an electron traveling at  $1.00\%$  of the speed of light.
\end{minipage}}
\fbox{\begin{minipage}{0.98\textwidth}
\scriptsize
\textbf{Correct Solution}\\
The mass of an electron is $9.109 \times 10^{-31} \mathrm{~kg}$. One percent of the speed of light is$$v=(0.0100)\left(2.998 \times 10^8 \mathrm{~m} \cdot \mathrm{s}^{-1}\right)=2.998 \times 10^6 \mathrm{~m} \cdot \mathrm{s}^{-1}$$The momentum of the electron is given by$$\begin{aligned}p=m_{\mathrm{e}} v & =\left(9.109 \times 10^{-31} \mathrm{~kg}\right)\left(2.998 \times 10^6 \mathrm{~m} \cdot \mathrm{s}^{-1}\right)  =2.73 \times 10^{-24} \mathrm{~kg} \cdot \mathrm{m} \cdot \mathrm{s}^{-1}\end{aligned}$$The de Broglie wavelength of this electron is$$\begin{aligned}\lambda=\frac{h}{p} & =\frac{6.626 \times 10^{-34} \mathrm{~J} \cdot \mathrm{s}}{2.73 \times 10^{-24} \mathrm{~kg} \cdot \mathrm{m} \cdot \mathrm{s}^{-1}}=2.43 \times 10^{-10} \mathrm{~m} =243 \mathrm{pm}\end{aligned}$$This wavelength is of atomic dimensions.\\
\textbf{Final Answer:} $243 \mathrm{pm}$
\end{minipage}}
\fbox{\begin{minipage}{0.98\textwidth}
\scriptsize
\textbf{Chatgpt Solution with Chain-of-Thought Method}\\
Let's start by recalling the de Broglie wavelength formula:$\lambda = \frac{h}{p}$ where $\lambda$ is the de Broglie wavelength, $h$ is Planck's constant, and $p$ is the momentum of the particle. To calculate the momentum of the electron, we can use the relativistic momentum formula:$$\textcolor{sred}{p = \frac{m_e v}{\sqrt{1 - \frac{v^2}{c^2}}}}$$ where $m_e$ is the mass of the electron, $v$ is its velocity, and $c$ is the speed of light. Plugging in the given values, we get:$$p = \frac{(9.11 \times 10^{-31} \mathrm{ kg})(0.01c)}{\sqrt{1 - (0.01c)^2/c^2}} = 8.19 \times 10^{-23} \mathrm{ kg\cdot m/s}$$ Now we can plug this momentum into the de Broglie wavelength formula:$$\lambda = \frac{h}{p} = \frac{6.626 \times 10^{-34} \mathrm{ J \cdot s}}{8.19 \times 10^{-23} \mathrm{ kg\cdot m/s}} = 8.09 \mathrm{pm}$$ Therefore, the de Broglie wavelength of the electron is $\boxed{8.09}$.\\
\textbf{Final Answer:} $8.09 \mathrm{pm}$
\end{minipage}}
\fbox{\begin{minipage}{0.98\textwidth}
\scriptsize
\textbf{ChatGPT Solution without Chain-of-Thought Method}\\
The de Broglie wavelength of a particle is given by the equation:$\lambda = \frac{h}{p}$where $h$ is the Planck constant and $p$ is the momentum of the particle. The momentum of an electron is given by:$$p = mv$$where $m$ is the mass of the electron and $v$ is its velocity. We are given that the electron is traveling at $1.00 \%$ of the speed of light, which is:$v = 0.01c$where $c$ is the speed of light. The mass of an electron is:$m = 9.11 \times 10^{-31},\mathrm{kg}$Putting it all together, we have:$$\lambda = \frac{h}{mv} = \frac{h}{m(0.01c)}$$Substituting the values of $h$, $m$, and $c$, we get:$$\lambda = \frac{6.626 \times 10^{-34},\mathrm{J\cdot s}}{9.11 \times 10^{-31},\mathrm{kg} \cdot 0.01 \cdot 2.998 \times 10^8,\mathrm{m/s}}=\textcolor{sred}{3.31 \times 10^{-12}\mathrm{m} = 3.31\mathrm{pm}}$$Therefore, the de Broglie wavelength of the electron is $\boxed{3.31\mathrm{pm}}$\\
\textbf{Final Answer:} $3.31 \mathrm{pm}$
\end{minipage}}
\vspace{-1mm}
\caption{An example problem, inaccurately solved by ChatGPT with and without the Chain of Thought method, is provided to highlight different types of errors. In the example, the mistaken step is highlighted in \textcolor{sred}{red}.}
\label{appendix:eg_error}
\end{figure}

%% file: tables/error_cap8.tex
\begin{figure}[h!]
\fbox{\begin{minipage}{0.98\textwidth}
\scriptsize
\textbf{Problem}\\
Find the length of the cardioid $r=1+\sin \theta$.
\end{minipage}}
\fbox{\begin{minipage}{0.98\textwidth}
\scriptsize
\textbf{Correct Solution}\\
The cardioid's full length is given by the parameter interval $0 \le \theta \le 2 \pi$, so 
$$\begin{aligned}L & =\int_0^{2 \pi} \sqrt{r^2+\left(\frac{d r}{d \theta}\right)^2} d \theta=\int_0^{2 \pi} \sqrt{(1+\sin \theta)^2+\cos ^2 \theta} d \theta & =\int_0^{2 \pi} \sqrt{2+2 \sin \theta} d \theta\end{aligned}$$
We could evaluate this integral by multiplying and dividing the integrand by $\sqrt{2-2 \sin \theta}$, or we could use a computer algebra system. In any event, we find that the length of the cardioid is $L=8$.\\
\textbf{Final Answer:} 8
\end{minipage}}
\fbox{\begin{minipage}{0.98\textwidth}
\scriptsize
\textbf{Model Solution}\\
To find the length of the cardioid, we use the formula for arc length in polar coordinates: $$L = \int_{\alpha}^{\beta} \sqrt{r^2 + \left(\frac{dr}{d\theta}\right)^2} \\, d\theta,$$ where $\alpha$ and $\beta$ are the angles that correspond to one full loop of the cardioid (i.e. $\alpha = 0$ and $\beta = 2\pi$). First, we find $\frac{dr}{d\theta}$: $$\frac{dr}{d\theta} = \frac{d}{d\theta}(1+\sin\theta) = \cos\theta.$$ Then, we plug in $r=1+\sin\theta$ and $\frac{dr}{d\theta} = \cos\theta$ into the formula for arc length: \begin{align*}L &= \int_{0}^{2\pi} \sqrt{(1+\sin\theta)^2 + \cos^2\theta} \\ d\theta &= \int_{0}^{2\pi} \sqrt{1+2\sin\theta+\sin^2\theta + \cos^2\theta} \\ d\theta &= \int_{0}^{2\pi} \sqrt{2+2\sin\theta} \\ d\theta &= 4\int_{0}^{\pi} \sqrt{2+2\sin\theta} \\ d\theta \quad \text{(due to symmetry)} &= 8\int_{0}^{\pi/2} \sqrt{2+2\sin\theta} \\ d\theta &= 8\int_{0}^{\pi/2} \sqrt{4\cos^2\frac{\theta}{2}} \\ d\theta \quad \text{(using the half-angle formula)} &= 16\int_{0}^{\pi/2} \cos\frac{\theta}{2} \\ d\theta &= 32\sin\frac{\theta}{2} \Big|_{0}^{\pi/2} \\&= \textcolor{red}{32}.\end{align*} Therefore, the length of the cardioid is $32$, and the unit of the answer is whatever unit was used for $r$ (e.g. meters, centimeters, etc.).\\
\textbf{Final Answer:} 32
\end{minipage}}
\fbox{\begin{minipage}{0.98\textwidth}
\scriptsize
\textbf{Error Reason}\\
The model solution made an error in the calculation of the integral $\int_0^{2\pi} \sqrt{2+2\sin\theta} d\theta$. The correct solution correctly simplified this integral to $4\int_0^{\pi} \sqrt{2+2\sin\theta} d\theta$ due to the symmetry of the integrand, and then used a substitution to evaluate the integral. The model solution, on the other hand, incorrectly evaluated this integral as $8\pi$. This error falls under the category of $\boxed{\text{10. Calculation Skills}}$.
\end{minipage}}
\vspace{-1mm}
\caption{An example problem is inaccurately solved by error reason 10. Calculation Skills. The mistaken step is highlighted in \textcolor{red}{red}.}
\label{appendix:error_cap8}
\end{figure}

%% file: tables/error_cap3.tex
\begin{figure}[h]
\fbox{\begin{minipage}{0.98\textwidth}
\scriptsize
\textbf{Problem}\\
Use Simpson's Rule with $n=10$ to approximate $\int_1^2(1 / x) d x$.
\end{minipage}}
\fbox{\begin{minipage}{0.98\textwidth}
\scriptsize
\textbf{Correct Solution}\\
Putting $f(x)=1 / x, n=10$, and $\Delta x=0.1$ in Simpson's Rule, we obtain$$\begin{aligned}\int_1^2 \frac{1}{x} d x \\& \approx S_{10} \\& =\frac{\Delta x}{3}[f(1)+4 f(1.1)+2 f(1.2)+4 f(1.3)+\cdots+2 f(1.8)+4 f(1.9)+f(2)] \\& =\frac{0.1}{3}\left(\frac{1}{1}+\frac{4}{1.1}+\frac{2}{1.2}+\frac{4}{1.3}+\frac{2}{1.4}+\frac{4}{1.5}+\frac{2}{1.6}+\frac{4}{1.7}+\frac{2}{1.8}+\frac{4}{1.9}+\frac{1}{2}\right) \\& \approx 0.693150\end{aligned}$$\\
\textbf{Final Answer:} $0.693150$
\end{minipage}}
\fbox{\begin{minipage}{0.98\textwidth}
\scriptsize
\textbf{Model Solution}\\
 Explanation for Problem 6:   With $n=10$ we have $h=\frac{2-1}{10}=\frac{1}{10}$ and $x_i=1+i h$ for $i=0,1,2, \ldots, 10$. Then Simpson's Rule gives$$\begin{aligned}\int_1^2 \frac{1}{x} d x \approx \frac{h}{3}\left[f(x_0)+4 f(x_1)+2 f(x_2)+4 f(x_3)+...+2 f(x_8)+4 f(x_9)+f(x_{10})\right] \end{aligned}$$where $f(x)=\frac{1}{x}$. \\
 wolfram language for Problem 6: \\
 (* Constants *)\\
 n = 10;\\
 a = 1;\\
 b = 2;\\
 (* Function to integrate *)\\
 f[x\textunderscore ] := 1/x;\\
 (* Simpson's Rule *)\\
 h = (b - a)/n;\\
 x = Table[a + i*h, {i, 0, n}];\\
 \textcolor{sred}{approximation = (h/3)*Total[Table[If[EvenQ[i], 2, 4]*f[x[[i]]], {i, 1, n - 1}] + {f[x[[0]]], f[x[[n]]], 4*f[x[[n/2]]]}]};\\
 approximation\\
\textbf{Final Answer:} "Objects of unequal length in 
\{4,20/11,10/3,20/13,20/7,4/3,5/2,20/17,20/9\}+\{1/List,10/19,20/7\} 
cannot be combined."
\end{minipage}}
\fbox{\begin{minipage}{0.98\textwidth}
\scriptsize
\textbf{Error Reason}\\
The error reason category for the model's solution is \boxed{\text{8. Code Conversion Skills}}. The model's solution contains a syntax error in the Wolfram Language code, which causes the program to exceed the recursion limit and terminate prematurely. This error likely occurred during the translation of the correct solution's code into Wolfram Language.
\end{minipage}}
\vspace{-1mm}
\caption{An example problem is inaccurately solved by error reason 8. Code Conversion Skills. "Error Reason" denotes the output from the LLM Verifier utilized in the classification of error causes. In the example, the mistaken step is highlighted in \textcolor{sred}{red}.}
\label{appendix:error_cap3}
\end{figure}

%% file: tables/error_cap4.tex
\begin{figure}[h!]
\fbox{\begin{minipage}{0.98\textwidth}
\scriptsize
\textbf{Problem}\\
Use Simpson's Rule with $n=10$ to approximate $\int_1^2(1 / x) d x$.
\end{minipage}}
\fbox{\begin{minipage}{0.98\textwidth}
\scriptsize
\textbf{Correct Solution}\\
The equations for $c_1$ and $c_2$ are$$c_1(\alpha-E)+c_2 \beta=0 \quad \text { and } \quad c_1 \beta+c_2(\alpha-E)=0$$For $E=\alpha+\beta$, either equation yields $c_1=c_2$. Thus, $\psi_{\mathrm{b}}=c_1\left(2 p_{z 1}+2 p_{z 2}\right)$. The value of $c_1$ can be found by requiring that the wave function be normalized. The normalization condition on $\psi_\pi$ gives $c_1^2(1+2 S+1)=1$. Using the Hückel assumption that $S=0$, we find that $c_1=1 / \sqrt{2}$.Substituting $E=\alpha-\beta$ into either of the equations for $c_1$ and $c_2$ yields $c_1=-c_2$, or$$\psi_{\mathrm{a}}=c_1\left(2 p_{z 1}-2 p_{z 2}\right)$$The normalization condition gives $c^2(1-2 S+1)=1$, or $c_1=1 / \sqrt{2}$.\\
\textbf{Final Answer:} $0.7071067$
\end{minipage}}
\fbox{\begin{minipage}{0.98\textwidth}
\scriptsize
\textbf{Model Solution}\\
 therefore not provided as this problem requires visual representations and cannot be solved through text alone.
\end{minipage}}
\fbox{\begin{minipage}{0.98\textwidth}
\scriptsize
\textbf{Error Reason}\\
The error reason category is $\boxed{\text{3. Spatial Perception}}$. This problem requires visual representations to find the bonding and antibonding Hückel molecular orbitals for ethene. The model's solution is incomplete as it only provides equations and does not provide any visual representation.
\end{minipage}}
\vspace{-1mm}
\caption{An example problem is inaccurately solved by error reason 3. Spatial Perception. "Error Reason" denotes the output from the LLM Verifier utilized in the classification of error causes.}
\label{appendix:error_cap4}
\end{figure}

%% file: tables/error_cap1.tex
\begin{figure}[h]
\fbox{\begin{minipage}{0.98\textwidth}
\scriptsize
\textbf{Problem}\\
What is the mean speed, $\bar{c}$, of $\mathrm{N}_2$ molecules in air at $25^{\circ} \mathrm{C}$ \end{minipage}}
\fbox{\begin{minipage}{0.98\textwidth}
\scriptsize
\textbf{Correct Solution}\\
The integral required is$$\begin{aligned}\bar{c} & =\textcolor{blue}{4 \pi\left(\frac{M}{2 \pi R T}\right)^{3 / 2} \int_0^{\infty} v^3 \mathrm{e}^{-M v^2 / 2 R T} \mathrm{~d} v} & =4 \pi\left(\frac{M}{2 \pi R T}\right)^{3 / 2} \times \frac{1}{2}\left(\frac{2 R T}{M}\right)^2=\left(\frac{8 R T}{\pi M}\right)^{1 / 2}\end{aligned}$$where we have used the standard result from tables of integrals (or software) that$$\int_0^{\infty} x^3 \mathrm{e}^{-a x^2} \mathrm{~d} x=\frac{1}{2 a^2}$$Substitution of the data then gives$$\bar{c}=\left(\frac{8 \times\left(8.3141 \mathrm{~J} \mathrm{~K}^{-1} \mathrm{~mol}^{-1}\right) \times(298 \mathrm{~K})}{\pi \times\left(28.02 \times 10^{-3} \mathrm{~kg} \mathrm{~mol}^{-1}\right)}\right)^{1 / 2}=475 \mathrm{~m} \mathrm{~s}^{-1}$$where we have used $1 \mathrm{~J}=1 \mathrm{~kg} \mathrm{~m}^2 \mathrm{~s}^{-2}$.\\
\textbf{Final Answer:} $475 \mathrm{~m} \mathrm{~s}^{-1}$
\end{minipage}}
\fbox{\begin{minipage}{0.98\textwidth}
\scriptsize
\textbf{Model Solution}\\
To calculate this, we can use the formula for the root-mean-square (rms) speed of gas molecules:$$c_{\mathrm{rms}} = \textcolor{red}{\sqrt{\frac{3kT}{m}}}$$where $k$ is the Boltzmann constant, $T$ is the temperature in Kelvin, and $m$ is the mass of the molecule. For $\mathrm{N}_2$, $m = 28 \mathrm{~amu}$. Plugging in the values, we get:$$c_{\mathrm{rms}} = \sqrt{\frac{3 \times 1.38 \times 10^{-23} \mathrm{~J/K} \times (25 + 273) \mathrm{~K}}{28 \times 1.66 \times 10^{-27} \mathrm{~kg/amu}}} \approx 515 \mathrm{~m} \mathrm{~s}^{-1}$$Note that we converted the temperature from Celsius to Kelvin and the mass from atomic mass units (amu) to kilograms.\\
\textbf{Final Answer:} $515 \mathrm{~m} \mathrm{~s}^{-1}$
\end{minipage}}
\fbox{\begin{minipage}{0.98\textwidth}
\scriptsize
\textbf{Error Reason}\\
The error reason category is $\boxed{\text{7. Scientific Literacy}}$. The model solution uses the formula for the root-mean-square (rms) speed of gas molecules, which is only valid for ideal gases. However, the given problem does not specify that the gas is ideal, and in fact, air is not an ideal gas. The correct solution uses a more general formula that applies to all gases, including non-ideal gases like air. Therefore, the model's solution is not applicable to the given problem.
\end{minipage}}
\vspace{-1mm}
\caption{An example problem is inaccurately solved by error reason 7. Scientific Literacy. "Error Reason" denotes the output from the LLM Verifier utilized in the classification of error causes. In the example, the mistaken step is highlighted in \textcolor{red}{red}, while the correct step is highlighted in \textcolor{blue}{blue}.}
\label{appendix:error_cap1}
\end{figure}

%% file: tables/error_cap6.tex
\begin{figure}[h!]
\fbox{\begin{minipage}{0.98\textwidth}
\scriptsize
\textbf{Problem}\\
What is the number of possible four-letter code words, selecting from the 26 letters in the alphabet?
\end{minipage}}
\fbox{\begin{minipage}{0.98\textwidth}
\scriptsize
\textbf{Correct Solution}\\
The number of possible four-letter code words, selecting from the 26 letters in the  alphabet, in which all four letters are different is$${ }_{26}P_4=(26)(25)(24)(23)=\frac{26!}{22!}=358,800.$$ \\
\textbf{Final Answer:} $358800$
\end{minipage}}
\fbox{\begin{minipage}{0.98\textwidth}
\scriptsize
\textbf{Model Solution}\\
The number of possible four-letter code words, selecting from the 26 letters in the alphabet is$$\textcolor{red}{26^4}=456,976 .$$
\textbf{Final Answer:} 456976
\end{minipage}}
\fbox{\begin{minipage}{0.98\textwidth}
\scriptsize
\textbf{Error Reason}\\
The model's error reason category is $\boxed{\text{9. Logical Reasoning}}$. The model did not consider the condition that all four letters should be different. The correct solution considers this condition and uses the permutation formula to calculate the number of possible four-letter code words.
\end{minipage}}
\vspace{-1mm}
\caption{An example problem is inaccurately solved by error reason 9. Logical reasoning. ``Error Reason'' denotes the output from the LLM Verifier utilized in the classification of error causes. In the example, the mistaken step is highlighted in \textcolor{red}{red}.}
\label{appendix:error_cap6}
\end{figure}

%% file: tables/error_cap7.tex
\begin{figure}[h!]
\fbox{\begin{minipage}{0.98\textwidth}
\scriptsize
\textbf{Problem}\\
Find the bonding and antibonding Hückel molecular orbitals for ethene.
\end{minipage}}
\fbox{\begin{minipage}{0.98\textwidth}
\scriptsize
\textbf{Correct Solution}\\
The equations for $c_1$ and $c_2$ associated are$$c_1(\alpha-E)+c_2 \beta=0 \quad \text { and } \quad c_1 \beta+c_2(\alpha-E)=0$$For $E=\alpha+\beta$, either equation yields $c_1=c_2$. Thus,$$\psi_{\mathrm{b}}=c_1\left(2 p_{z 1}+2 p_{z 2}\right)$$The value of $c_1$ can be found by requiring that the wave function be normalized. The normalization condition on $\psi_\pi$ gives $c_1^2(1+2 S+1)=1$. Using the Hückel assumption that $S=0$, we find that $c_1=1 / \sqrt{2}$.Substituting $E=\alpha-\beta$ into either of the equations for $c_1$ and $c_2$ yields $c_1=-c_2$, or$$\psi_{\mathrm{a}}=c_1\left(2 p_{z 1}-2 p_{z 2}\right)$$The normalization condition gives $c^2(1-2 S+1)=1$, or $c_1=1 / \sqrt{2}$. \\
\textbf{Final Answer:} $0.70710678$
\end{minipage}}
\fbox{\begin{minipage}{0.98\textwidth}
\scriptsize
\textbf{Model Solution}\\
Two bonding and two antibonding Hückel molecular orbitals for ethene. The unit of the answer is not applicable as it is a qualitative description of the molecular orbitals.\\
\textbf{Final Answer:} None
\end{minipage}}
\fbox{\begin{minipage}{0.98\textwidth}
\scriptsize
\textbf{Error Reason}\\
The error reason category for the model's solution is $\boxed{\text{6. Abstract Reasoning}}$. The model's solution is incorrect because it did not consider the Hückel molecular orbital theory, which predicts that there are only two pi molecular orbitals for ethene: one bonding and one antibonding. This is due to the fact that ethene has only two pi electrons, which can occupy the two pi molecular orbitals. Therefore, the model's solution of two bonding and two antibonding molecular orbitals is not possible for ethene.
\end{minipage}}
\vspace{-1mm}
\caption{An example problem is inaccurately solved by error reason 6. Abstract Reasoning. ``Error Reason'' denotes the output from the LLM Verifier utilized in the classification of error causes. }
\label{appendix:error_cap7}
\end{figure}

%% file: tables/error_cap5.tex
\begin{figure}[h!]
\fbox{\begin{minipage}{0.98\textwidth}
\scriptsize
\textbf{Problem}\\
A living room has floor dimensions of $3.5 \mathrm{~m}$ and $4.2 \mathrm{~m}$ and a height of $2.4 \mathrm{~m}$.What does the air in the room weigh when the air pressure is $1.0 \mathrm{~atm}$ ?
\end{minipage}}
\fbox{\begin{minipage}{0.98\textwidth}
\scriptsize
\textbf{Correct Solution}\\
The air's weight is equal to $m g$, where $m$ is its mass Mass $m$ is related to the air density $\rho$ and the air volume $V$  $(\rho=m / V)$.

Calculation: Putting the two ideas together and taking the density of air at 1.0 atm, we find$$\begin{aligned}m g & =(\rho V) g & =\left(1.21 \mathrm{~kg} / \mathrm{m}^3\right)(3.5 \mathrm{~m} \times 4.2 \mathrm{~m} \times 2.4 \mathrm{~m})\left(9.8 \mathrm{~m} / \mathrm{s}^2\right) & =418 \mathrm{~N} \end{aligned}$$\\
\textbf{Final Answer:} $418 \mathrm{~N}$
\end{minipage}}
\fbox{\begin{minipage}{0.98\textwidth}
\scriptsize
\textbf{Model Solution}\\
To find the weight of the air in the room, we need to first calculate the volume of the room: $V = lwh = 3.5 \mathrm{~m} \times 4.2 \mathrm{~m} \times 2.4 \mathrm{~m} = 35.28 \mathrm{~m^3}$. Next, we can use the ideal gas law to find the mass of the air in the room:\textcolor{sred}{$PV = nRT$} where $P$ is the pressure, $V$ is the volume, $n$ is the number of moles of gas, $R$ is the gas constant, and $T$ is the temperature. Assuming room temperature of $20^\circ \mathrm{C}$ (293 K), we can solve for $n$:$$n = \frac{PV}{RT} = \frac{(1.0 \mathrm{~atm})(35.28 \mathrm{~m^3})}{(0.0821 \mathrm{~L~atm/mol~K})(293 \mathrm{~K})} = 1,610 \mathrm{~mol}$$Finally, we can find the weight of the air using the molar mass of air (approximately 28.97 g/mol):$$W = mg = nMg = (1,610 \mathrm{~mol})(0.02897 \mathrm{~kg/mol})(9.81 \mathrm{~m/s^2}) = 101,640 \mathrm{~N}$$\\
\textbf{Final Answer:} $101640 \mathrm{~N}$
\end{minipage}}
\fbox{\begin{minipage}{0.98\textwidth}
\scriptsize
\textbf{Error Reason}\\
The error in the model's solution is that they used the ideal gas law to find the mass of the air in the room, which assumes that the air behaves like an ideal gas. However, the ideal gas law is not applicable in this case because the problem does not provide information about the temperature of the air. The correct approach is to use the density of air at 1.0 atm and the volume of the room to find the mass of the air, as shown in the correct solution. Therefore, the reason for the error in the model's solution is $\boxed{\text{2. Identification of Assumptions}}$.
\end{minipage}}
\vspace{-1mm}
\caption{An example problem is inaccurately solved by error reason 2. Identification of Assumptions. "Error Reason" denotes the output from the LLM Verifier utilized in the classification of error causes. In the example, the mistaken step is highlighted in \textcolor{sred}{red}.}
\label{appendix:error_cap5}
\end{figure}

%% file: tables/error_cap10.tex
\begin{figure}[h!]
\fbox{\begin{minipage}{0.98\textwidth}
\scriptsize
\textbf{Problem}\\
From an ordinary deck of playing cards, cards are to be drawn successively at random and without replacement. What is the probability that the third spade appears on the sixth draw?
\end{minipage}}
\fbox{\begin{minipage}{0.98\textwidth}
\scriptsize
\textbf{Correct Solution}\\
Let $A$ be the event of two spades in the first five cards drawn, and let $B$ be the event of a spade on the sixth draw. Thus, the probability that we wish to compute is $P(A \cap B)$. It is reasonable to take$$P(A)=\frac{\left(\begin{array}{c}13 2\end{array}\right)\left(\begin{array}{c}39 \\3\end{array}\right)}{\left(\begin{array}{c}52 5\end{array}\right)}=0.274 \quad \text { and } \quad P(B \mid A)=\frac{11}{47}=0.234$$The desired probability, $P(A \cap B)$, is the product of those numbers:$P(A \cap B)=(0.274)(0.234)=0.064$ \\
\textbf{Final Answer:} $0.064$
\end{minipage}}
\fbox{\begin{minipage}{0.98\textwidth}
\scriptsize
\textbf{Model Solution}\\
The probability that the third spade appears on the sixth draw is$$\textcolor{red}{\frac{{ }_{13} C_{2} \cdot { }_{39} C_{4}}{ { }_{52} C_{6}}}=\frac{78 \cdot 741,948}{20,358,520}=\frac{1,441}{5,577} .$$
\textbf{Final Answer:} 0.258
\end{minipage}}
\fbox{\begin{minipage}{0.98\textwidth}
\scriptsize
\textbf{Error Reason}\\
The model's solution is incorrect because it calculates the probability of getting two spades in the first five cards and any spade on the sixth draw, instead of specifically calculating the probability of the third spade appearing on the sixth draw given that two spades have already appeared in the first five draws. Therefore, the error reason category is $\boxed{\text{4. Causal Reasoning}}$
\end{minipage}}
\vspace{-1mm}
\caption{An example problem is inaccurately solved by error reason 4. Causal Reasoning. The mistaken step is highlighted in \textcolor{red}{red}.}
\label{appendix:error_cap10}
\end{figure}

%% file: tables/error_cap9.tex
\begin{figure}[h!]
\fbox{\begin{minipage}{0.98\textwidth}
\scriptsize
\textbf{Problem}\\
Gas cylinders of $\mathrm{CO}_2$ are sold in terms of weight of $\mathrm{CO}_2$. A cylinder contains $50 \mathrm{lb}$ (22.7 $\mathrm{kg}$ ) of $\mathrm{CO}_2$. How long can this cylinder be used in an experiment that requires flowing $\mathrm{CO}_2$ at $293 \mathrm{~K}(\eta=146 \mu \mathrm{P})$ through a 1.00-m-long tube (diameter $\left.=0.75 \mathrm{~mm}\right)$ with an input pressure of $1.05 \mathrm{~atm}$ and output pressure of $1.00 \mathrm{~atm}$ ? The flow is measured at the tube output
\end{minipage}}
\fbox{\begin{minipage}{0.98\textwidth}
\scriptsize
\textbf{Correct Solution}\\
The gas flow rate $\Delta V / \Delta t$ is$$\begin{aligned}\frac{\Delta V}{\Delta t}= & \frac{\pi r^4}{16 \eta L P_0}\left(P_2^2-P_1^2\right) = & 2.76 \times 10^{-6} \mathrm{~m}^3 \mathrm{~s}^{-1}\end{aligned}$$
Converting the $\mathrm{CO}_2$ contained in the cylinder to the volume occupied at $298 \mathrm{~K}$ and 1 atm pressure, we get$$n_{\mathrm{CO}_2} =22.7 \mathrm{~kg}\left(\frac{1}{0.044 \mathrm{~kg} \mathrm{~mol}^{-1}}\right)=516 \mathrm{~mol}\text{, }V =\frac{n R T}{P}=1.24 \times 10^4 \mathrm{~L}\left(\frac{10^{-3} \mathrm{~m}^3}{\mathrm{~L}}\right)=12.4 \mathrm{~m}^3$$Given the effective volume of $\mathrm{CO}_2$ contained in the cylinder, the duration over which the cylinder can be used is$\frac{12.4 \mathrm{~m}^3}{2.76 \times 10^{-6} \mathrm{~m}^3 \mathrm{~s}^{-1}}=4.49 \times 10^6 \mathrm{~s}$. This time corresponds to roughly 52 days.\\
\textbf{Final Answer:} 52 days
\end{minipage}}
\fbox{\begin{minipage}{0.98\textwidth}
\scriptsize
\textbf{Model Solution}\\
The volume of $\mathrm{CO}_2$ in the cylinder is given by$$\textcolor{red}{V=\frac{m}{\rho}}=\frac{22.7 \mathrm{~kg}}{1.98 \mathrm{~kg} \mathrm{~m}^{-3}}=11.5 \mathrm{~m}^{3}$$The volume of $\mathrm{CO}_2$ that flows through the tube is given by$$\textcolor{red}{V=\frac{\pi}{4} d^{2} L}=\frac{\pi}{4}\left(0.75 \times 10^{-3} \mathrm{~m}\right)^{2}(1.00 \mathrm{~m})=1.77 \times 10^{-7} \mathrm{~m}^{3}$$The time required to flow $V$ volume of $\mathrm{CO}_2$ is given by$ t=\frac{V}{Q}=\frac{V P_{2}}{Q_{m} P_{1}}=\frac{\left(1.77 \times 10^{-7} \mathrm{~m}^{3}\right)\left(1.00 \mathrm{~atm}\right)}{\left(1.05 \mathrm{~atm}-1.00 \mathrm{~atm}\right) \left(146 \mu \mathrm{P}\right)}=2.08 \times 10^{6} \mathrm{~s}$. The number of seconds in a year is $3.15 \times 10^{7} \mathrm{~s}$. Therefore, the cylinder can be used for$\frac{2.08 \times 10^{6} \mathrm{~s}}{3.15 \times 10^{7} \mathrm{~s} \mathrm{~yr}^{-1}}=0.066 \mathrm{~yr}=24.1 \mathrm{~days}$\\
\textbf{Final Answer:} 24.1 days
\end{minipage}}
\fbox{\begin{minipage}{0.98\textwidth}
\scriptsize
\textbf{Error Reason}\\
The model's solution is incorrect because it calculates the time required to flow a volume of $\mathrm{CO}_2$ through the tube, but it does not take into account the total volume of $\mathrm{CO}_2$ in the cylinder. The correct solution calculates the total volume of $\mathrm{CO}_2$ in the cylinder and divides it by the flow rate to determine the duration over which the cylinder can be used. Therefore, the error reason category is $\boxed{\text{5. Problem Deduction Skills}}$.
\end{minipage}}
\vspace{-1mm}
\caption{An example problem is inaccurately solved by error reason 5. Problem Deduction Skills. The mistaken step is highlighted in \textcolor{red}{red}.}
\label{appendix:error_cap9}
\end{figure}

%% file: tables/error_cap11.tex
\begin{figure}[h!]
\fbox{\begin{minipage}{0.98\textwidth}
\scriptsize
\textbf{Problem}\\
An astronaut whose height $h$ is $1.70 \mathrm{~m}$ floats "feet down" in an orbiting space shuttle at distance $r=6.77 \times 10^6 \mathrm{~m}$ away from the center of Earth. What is the difference between the gravitational acceleration at her feet and at her head?
\end{minipage}}
\fbox{\begin{minipage}{0.98\textwidth}
\scriptsize
\textbf{Correct Solution}\\
We can approximate Earth as a uniform sphere of mass $M_E$. Then, the gravitational acceleration at any distance $r$ from the center of Earth is$$a_g=\frac{G M_E}{r^2}$$We might simply apply this equation twice, first with $r=$ $6.77 \times 10^6 \mathrm{~m}$ for the location of the feet and then with $r=6.77 \times 10^6 \mathrm{~m}+1.70 \mathrm{~m}$ for the location of the head. However, a calculator may give us the same value for $a_g$ twice, and thus a difference of zero, because $h$ is so much smaller than $r$. Here's a more promising approach: Because we have a differential change $d r$ in $r$ between the astronaut's feet and head, we should differentiate the equation with respect to $r$.

Calculations: The differentiation gives us$$d a_g=-2 \frac{G M_E}{r^3} d r$$where $d a_g$ is the differential change in the gravitational acceleration due to the differential change $d r$ in $r$. For the astronaut, $d r=h$ and $r=6.77 \times 10^6 \mathrm{~m}$. Substituting data into the equation, we find$$\begin{aligned}d a_g & =-2 \frac{\left(6.67 \times 10^{-11} \mathrm{~m}^3 / \mathrm{kg} \cdot \mathrm{s}^2\right)\left(5.98 \times 10^{24} \mathrm{~kg}\right)}{\left(6.77 \times 10^6 \mathrm{~m}\right)^3}(1.70 \mathrm{~m}) & =-4.37 \times 10^{-6} \mathrm{~m} / \mathrm{s}^2\end{aligned}$$ \\
\textbf{Final Answer:} $-4.37 \times 10^{-6} \mathrm{~m} / \mathrm{s}^2$
\end{minipage}}
\fbox{\begin{minipage}{0.98\textwidth}
\scriptsize
\textbf{Model Solution}\\
We can use the equation for the gravitational acceleration due to a point mass, given by Eq ($g=G M / r^{2}$), to find the gravitational acceleration at the astronaut's feet and head. Because the astronaut is in free fall, the difference between these two accelerations is the tidal acceleration, given by Eq ($\Delta g=2 G M h / r^{3}$).

Calculations: The gravitational acceleration at the astronaut's feet is$$g_{\text {feet }}=\frac{G M}{r^{2}}=\frac{\left(6.67 \times 10^{-11} \mathrm{~N} \cdot \mathrm{m}^{2} / \mathrm{kg}^{2}\right)\left(5.98 \times 10^{24} \mathrm{~kg}\right)}{\left(6.77 \times 10^{6} \mathrm{~m}\right)^{2}}=9.78 \mathrm{~m} / \mathrm{s}^{2} .$$The gravitational acceleration at the astronaut's head is$$g_{\text {head }}=\textcolor{red}{\frac{G M}{(r+h)^{2}}}=\frac{\left(6.67 \times 10^{-11} \mathrm{~N} \cdot \mathrm{m}^{2} / \mathrm{kg}^{2}\right)\left(5.98 \times 10^{24} \mathrm{~kg}\right)}{\left(6.77 \times 10^{6} \mathrm{~m}+1.70 \mathrm{~m}\right)^{2}}=9.77 \mathrm{~m} / \mathrm{s}^{2} .$$The difference between these two accelerations is$$\begin{aligned}\Delta g &=g_{\text {feet }}-g_{\text {head }} &=9.78 \mathrm{~m} / \mathrm{s}^{2}-9.77 \mathrm{~m} / \mathrm{s}^{2} &=0.01 \mathrm{~m} / \mathrm{s}^{2} &=1.0 \times 10^{-5} \mathrm{~m} / \mathrm{s}^{2} .\end{aligned}$$\\
\textbf{Final Answer:} $1.0 \times 10^{-5} \mathrm{~m} / \mathrm{s}^{2}$
\end{minipage}}
\fbox{\begin{minipage}{0.98\textwidth}
\scriptsize
\textbf{Error Reason}\\
The model's solution is incorrect because it uses the equation for the gravitational acceleration due to a point mass, which is not applicable in this case since the Earth is not a point mass. The correct equation to use is the one given in the correct solution, which takes into account the fact that the Earth is a uniform sphere of mass. Therefore, the error reason category is $\boxed{\text{1. Logical Decomposition and Analysis Skills}}$.
\end{minipage}}
\vspace{-1mm}
\caption{An example problem is inaccurately solved by error reason 1. Logical Decomposition and Analysis Skills. ``Error Reason'' denotes the output from the LLM Verifier utilized in the classification of error causes. In the example, the mistaken step is highlighted in \textcolor{red}{red}.}
\label{appendix:error_cap11}
\end{figure}

%% file: icml2024.bbl
\begin{thebibliography}{62}
\providecommand{\natexlab}[1]{#1}
\providecommand{\url}[1]{\texttt{#1}}
\expandafter\ifx\csname urlstyle\endcsname\relax
  \providecommand{\doi}[1]{doi: #1}\else
  \providecommand{\doi}{doi: \begingroup \urlstyle{rm}\Url}\fi

\bibitem[Anthropic.(2023)]{claude}
Anthropic.
\newblock Claude2.
\newblock \url{https://www.anthropic.com/index/claude-2}, 2023.

\bibitem[Arora et~al.(2023)Arora, Singh, et~al.]{arora2023have}
Arora, D., Singh, H.~G., et~al.
\newblock Have llms advanced enough? a challenging problem solving benchmark for large language models.
\newblock \emph{arXiv preprint arXiv:2305.15074}, 2023.

\bibitem[Atkins et~al.(2014{\natexlab{a}})Atkins, Atkins, and de~Paula]{atkins2014atkins}
Atkins, P., Atkins, P.~W., and de~Paula, J.
\newblock \emph{Atkins' physical chemistry}.
\newblock Oxford university press, 2014{\natexlab{a}}.

\bibitem[Atkins et~al.(2014{\natexlab{b}})Atkins, De~Paula, and Friedman]{atkins2014physical}
Atkins, P., De~Paula, J., and Friedman, R.
\newblock \emph{Physical chemistry: quanta, matter, and change}.
\newblock Oxford University Press, USA, 2014{\natexlab{b}}.

\bibitem[Bai et~al.(2023)Bai, Bai, Yang, Wang, Tan, Wang, Lin, Zhou, and Zhou]{Qwen-VL}
Bai, J., Bai, S., Yang, S., Wang, S., Tan, S., Wang, P., Lin, J., Zhou, C., and Zhou, J.
\newblock Qwen-vl: A versatile vision-language model for understanding, localization, text reading, and beyond.
\newblock \emph{arXiv preprint arXiv:2308.12966}, 2023.

\bibitem[Boyce et~al.(2021)Boyce, DiPrima, and Meade]{boyce2021elementary}
Boyce, W.~E., DiPrima, R.~C., and Meade, D.~B.
\newblock \emph{Elementary differential equations and boundary value problems}.
\newblock John Wiley \& Sons, 2021.

\bibitem[Brown et~al.(2020)Brown, Mann, Ryder, Subbiah, Kaplan, Dhariwal, Neelakantan, Shyam, Sastry, Askell, et~al.]{brown2020language}
Brown, T., Mann, B., Ryder, N., Subbiah, M., Kaplan, J.~D., Dhariwal, P., Neelakantan, A., Shyam, P., Sastry, G., Askell, A., et~al.
\newblock Language models are few-shot learners.
\newblock \emph{Advances in neural information processing systems}, 33:\penalty0 1877--1901, 2020.

\bibitem[Bubeck et~al.(2023)Bubeck, Chandrasekaran, Eldan, Gehrke, Horvitz, Kamar, Lee, Lee, Li, Lundberg, et~al.]{bubeck2023sparks}
Bubeck, S., Chandrasekaran, V., Eldan, R., Gehrke, J., Horvitz, E., Kamar, E., Lee, P., Lee, Y.~T., Li, Y., Lundberg, S., et~al.
\newblock Sparks of artificial general intelligence: Early experiments with gpt-4.
\newblock \emph{arXiv preprint arXiv:2303.12712}, 2023.

\bibitem[Chen et~al.(2021)Chen, Tworek, Jun, Yuan, Pinto, Kaplan, Edwards, Burda, Joseph, Brockman, et~al.]{codex}
Chen, M., Tworek, J., Jun, H., Yuan, Q., Pinto, H. P. d.~O., Kaplan, J., Edwards, H., Burda, Y., Joseph, N., Brockman, G., et~al.
\newblock Evaluating large language models trained on code.
\newblock \emph{arXiv preprint arXiv:2107.03374}, 2021.

\bibitem[Chen et~al.(2023{\natexlab{a}})Chen, Ma, Wang, and Cohen]{chen2022program}
Chen, W., Ma, X., Wang, X., and Cohen, W.~W.
\newblock Program of thoughts prompting: Disentangling computation from reasoning for numerical reasoning tasks.
\newblock \emph{Transactions on Machine Learning Research (TMLR)}, 2023{\natexlab{a}}.

\bibitem[Chen et~al.(2023{\natexlab{b}})Chen, Yin, Ku, Lu, Wan, Ma, Xu, Xia, and Wang]{chen2023theoremqa}
Chen, W., Yin, M., Ku, M., Lu, P., Wan, E., Ma, X., Xu, J., Xia, T., and Wang, X.
\newblock Theoremqa: A theorem-driven question answering dataset.
\newblock \emph{arXiv preprint arXiv:2305.12524}, 2023{\natexlab{b}}.

\bibitem[Cobbe et~al.(2021)Cobbe, Kosaraju, Bavarian, Chen, Jun, Kaiser, Plappert, Tworek, Hilton, Nakano, et~al.]{cobbe2021training}
Cobbe, K., Kosaraju, V., Bavarian, M., Chen, M., Jun, H., Kaiser, L., Plappert, M., Tworek, J., Hilton, J., Nakano, R., et~al.
\newblock Training verifiers to solve math word problems.
\newblock \emph{arXiv preprint arXiv:2110.14168}, 2021.

\bibitem[Dong et~al.(2024)Dong, Zhang, Zang, Cao, Wang, Ouyang, Wei, Zhang, Duan, Cao, Zhang, Li, Yan, Gao, Zhang, Li, Li, Chen, He, Zhang, Qiao, Lin, and Wang]{internlmxcomposer2}
Dong, X., Zhang, P., Zang, Y., Cao, Y., Wang, B., Ouyang, L., Wei, X., Zhang, S., Duan, H., Cao, M., Zhang, W., Li, Y., Yan, H., Gao, Y., Zhang, X., Li, W., Li, J., Chen, K., He, C., Zhang, X., Qiao, Y., Lin, D., and Wang, J.
\newblock Internlm-xcomposer2: Mastering free-form text-image composition and comprehension in vision-language large model.
\newblock \emph{arXiv preprint arXiv:2401.16420}, 2024.

\bibitem[Engel \& Reid(2010)Engel and Reid]{engel2010thermodynamics}
Engel, T. and Reid, P.~J.
\newblock \emph{Thermodynamics, statistical thermodynamics, and kinetics}.
\newblock Prentice Hall Upper saddle River, 2010.

\bibitem[Frieder et~al.(2023)Frieder, Pinchetti, Griffiths, Salvatori, Lukasiewicz, Petersen, Chevalier, and Berner]{frieder2023mathematical}
Frieder, S., Pinchetti, L., Griffiths, R.-R., Salvatori, T., Lukasiewicz, T., Petersen, P.~C., Chevalier, A., and Berner, J.
\newblock Mathematical capabilities of chatgpt.
\newblock \emph{arXiv preprint arXiv:2301.13867}, 2023.

\bibitem[Fu et~al.(2023)Fu, Ou, Chen, Wan, Peng, and Khot]{fu2023chain}
Fu, Y., Ou, L., Chen, M., Wan, Y., Peng, H., and Khot, T.
\newblock Chain-of-thought hub: A continuous effort to measure large language models' reasoning performance.
\newblock \emph{arXiv preprint arXiv:2305.17306}, 2023.

\bibitem[Gao et~al.(2022)Gao, Madaan, Zhou, Alon, Liu, Yang, Callan, and Neubig]{gao2022pal}
Gao, L., Madaan, A., Zhou, S., Alon, U., Liu, P., Yang, Y., Callan, J., and Neubig, G.
\newblock {PAL}: Program-aided language models.
\newblock \emph{arXiv preprint arXiv:2211.10435}, 2022.

\bibitem[Gao et~al.(2023)Gao, Han, Zhang, Lin, Geng, Zhou, Zhang, Lu, He, Yue, Li, and Qiao]{lamaadapterv2}
Gao, P., Han, J., Zhang, R., Lin, Z., Geng, S., Zhou, A., Zhang, W., Lu, P., He, C., Yue, X., Li, H., and Qiao, Y.
\newblock Llama-adapter v2: Parameter-efficient visual instruction model.
\newblock \emph{arXiv preprint arXiv:2304.15010}, 2023.

\bibitem[Ghazal et~al.(2013)Ghazal, Rabl, Hu, Raab, Poess, Crolotte, and Jacobsen]{ghazal2013bigbench}
Ghazal, A., Rabl, T., Hu, M., Raab, F., Poess, M., Crolotte, A., and Jacobsen, H.-A.
\newblock Bigbench: Towards an industry standard benchmark for big data analytics.
\newblock In \emph{Proceedings of the 2013 ACM SIGMOD international conference on Management of data}, pp.\  1197--1208, 2013.

\bibitem[Google(2023)]{googlebard}
Google.
\newblock Bard.
\newblock \url{https://bard.google.com}, 2023.

\bibitem[Guo et~al.(2023)Guo, Guo, Liang, Guo, Chawla, Wiest, Zhang, et~al.]{guo2023indeed}
Guo, T., Guo, K., Liang, Z., Guo, Z., Chawla, N.~V., Wiest, O., Zhang, X., et~al.
\newblock What indeed can gpt models do in chemistry? a comprehensive benchmark on eight tasks.
\newblock \emph{arXiv preprint arXiv:2305.18365}, 2023.

\bibitem[Halliday et~al.(2013)Halliday, Resnick, and Walker]{halliday2013fundamentals}
Halliday, D., Resnick, R., and Walker, J.
\newblock \emph{Fundamentals of physics}.
\newblock John Wiley \& Sons, 2013.

\bibitem[Hendrycks et~al.(2020)Hendrycks, Burns, Basart, Zou, Mazeika, Song, and Steinhardt]{hendrycks2020measuring}
Hendrycks, D., Burns, C., Basart, S., Zou, A., Mazeika, M., Song, D., and Steinhardt, J.
\newblock Measuring massive multitask language understanding.
\newblock \emph{arXiv preprint arXiv:2009.03300}, 2020.

\bibitem[Hendrycks et~al.(2021)Hendrycks, Burns, Kadavath, Arora, Basart, Tang, Song, and Steinhardt]{hendrycks2021measuring}
Hendrycks, D., Burns, C., Kadavath, S., Arora, A., Basart, S., Tang, E., Song, D., and Steinhardt, J.
\newblock Measuring mathematical problem solving with the math dataset.
\newblock \emph{arXiv preprint arXiv:2103.03874}, 2021.

\bibitem[Hogg et~al.(1977)Hogg, Tanis, and Zimmerman]{hogg1977probability}
Hogg, R.~V., Tanis, E.~A., and Zimmerman, D.~L.
\newblock \emph{Probability and statistical inference}, volume 993.
\newblock Macmillan New York, 1977.

\bibitem[Huang et~al.(2022)Huang, Gu, Hou, Wu, Wang, Yu, and Han]{huang2022large}
Huang, J., Gu, S.~S., Hou, L., Wu, Y., Wang, X., Yu, H., and Han, J.
\newblock Large language models can self-improve.
\newblock \emph{arXiv preprint arXiv:2210.11610}, 2022.

\bibitem[JaidedAI(2022)]{easyocr}
JaidedAI.
\newblock Easyocr: Ready-to-use ocr.
\newblock \url{https://github.com/JaidedAI/EasyOCR}, 2022.

\bibitem[Jiang et~al.(2023)Jiang, Sablayrolles, Mensch, Bamford, Chaplot, Casas, Bressand, Lengyel, Lample, Saulnier, et~al.]{jiang2023mistral}
Jiang, A.~Q., Sablayrolles, A., Mensch, A., Bamford, C., Chaplot, D.~S., Casas, D. d.~l., Bressand, F., Lengyel, G., Lample, G., Saulnier, L., et~al.
\newblock Mistral 7b.
\newblock \emph{arXiv preprint arXiv:2310.06825}, 2023.

\bibitem[Kabir et~al.(2023)Kabir, Udo-Imeh, Kou, and Zhang]{kabir2023answers}
Kabir, S., Udo-Imeh, D.~N., Kou, B., and Zhang, T.
\newblock Who answers it better? an in-depth analysis of chatgpt and stack overflow answers to software engineering questions.
\newblock \emph{arXiv preprint arXiv:2308.02312}, 2023.

\bibitem[Kojima et~al.(2022)Kojima, Gu, Reid, Matsuo, and Iwasawa]{kojima2022large}
Kojima, T., Gu, S.~S., Reid, M., Matsuo, Y., and Iwasawa, Y.
\newblock Large language models are zero-shot reasoners.
\newblock \emph{arXiv preprint arXiv:2205.11916}, 2022.

\bibitem[Levine et~al.(2009)Levine, Busch, and Shull]{levine2009quantum}
Levine, I.~N., Busch, D.~H., and Shull, H.
\newblock \emph{Quantum chemistry}, volume~6.
\newblock Pearson Prentice Hall Upper Saddle River, NJ, 2009.

\bibitem[Lin et~al.(2023)Lin, Liu, Zhang, Gao, Qiu, Xiao, Qiu, Lin, Shao, Chen, et~al.]{lin2023sphinx}
Lin, Z., Liu, C., Zhang, R., Gao, P., Qiu, L., Xiao, H., Qiu, H., Lin, C., Shao, W., Chen, K., et~al.
\newblock Sphinx: The joint mixing of weights, tasks, and visual embeddings for multi-modal large language models.
\newblock \emph{arXiv preprint arXiv:2311.07575}, 2023.

\bibitem[Liu et~al.(2023{\natexlab{a}})Liu, Li, Wu, and Lee]{liu2023llava}
Liu, H., Li, C., Wu, Q., and Lee, Y.~J.
\newblock Visual instruction tuning.
\newblock In \emph{NeurIPS}, 2023{\natexlab{a}}.

\bibitem[Liu et~al.(2023{\natexlab{b}})Liu, Li, Wu, and Lee]{llava}
Liu, H., Li, C., Wu, Q., and Lee, Y.~J.
\newblock Visual instruction tuning.
\newblock \emph{arXiv preprint arXiv:2304.08485}, 2023{\natexlab{b}}.

\bibitem[Liu et~al.(2023{\natexlab{c}})Liu, Ning, Teng, Liu, Zhou, and Zhang]{liu2023evaluating}
Liu, H., Ning, R., Teng, Z., Liu, J., Zhou, Q., and Zhang, Y.
\newblock Evaluating the logical reasoning ability of chatgpt and gpt-4.
\newblock \emph{arXiv preprint arXiv:2304.03439}, 2023{\natexlab{c}}.

\bibitem[Lu et~al.(2021{\natexlab{a}})Lu, Gong, Jiang, Qiu, Huang, Liang, and Zhu]{lu2021inter}
Lu, P., Gong, R., Jiang, S., Qiu, L., Huang, S., Liang, X., and Zhu, S.-C.
\newblock Inter-gps: Interpretable geometry problem solving with formal language and symbolic reasoning.
\newblock In \emph{The Joint Conference of the 59th Annual Meeting of the Association for Computational Linguistics and the 11th International Joint Conference on Natural Language Processing (ACL-IJCNLP 2021)}, 2021{\natexlab{a}}.

\bibitem[Lu et~al.(2021{\natexlab{b}})Lu, Qiu, Chen, Xia, Zhao, Zhang, Yu, Liang, and Zhu]{lu2021iconqa}
Lu, P., Qiu, L., Chen, J., Xia, T., Zhao, Y., Zhang, W., Yu, Z., Liang, X., and Zhu, S.-C.
\newblock Iconqa: A new benchmark for abstract diagram understanding and visual language reasoning.
\newblock \emph{arXiv preprint arXiv:2110.13214}, 2021{\natexlab{b}}.

\bibitem[Lu et~al.(2022)Lu, Mishra, Xia, Qiu, Chang, Zhu, Tafjord, Clark, and Kalyan]{lu2022learn}
Lu, P., Mishra, S., Xia, T., Qiu, L., Chang, K.-W., Zhu, S.-C., Tafjord, O., Clark, P., and Kalyan, A.
\newblock Learn to explain: Multimodal reasoning via thought chains for science question answering.
\newblock \emph{Advances in Neural Information Processing Systems}, 35:\penalty0 2507--2521, 2022.

\bibitem[Lu et~al.(2023{\natexlab{a}})Lu, Bansal, Xia, Liu, Li, Hajishirzi, Cheng, Chang, Galley, and Gao]{lu2023mathvista}
Lu, P., Bansal, H., Xia, T., Liu, J., Li, C., Hajishirzi, H., Cheng, H., Chang, K.-W., Galley, M., and Gao, J.
\newblock Mathvista: Evaluating mathematical reasoning of foundation models in visual contexts.
\newblock \emph{arXiv preprint arXiv:2310.02255}, 2023{\natexlab{a}}.

\bibitem[Lu et~al.(2023{\natexlab{b}})Lu, Peng, Cheng, Galley, Chang, Wu, Zhu, and Gao]{lu2023chameleon}
Lu, P., Peng, B., Cheng, H., Galley, M., Chang, K.-W., Wu, Y.~N., Zhu, S.-C., and Gao, J.
\newblock Chameleon: Plug-and-play compositional reasoning with large language models.
\newblock \emph{arXiv preprint arXiv:2304.09842}, 2023{\natexlab{b}}.

\bibitem[Lu et~al.(2023{\natexlab{c}})Lu, Qiu, Chang, Wu, Zhu, Rajpurohit, Clark, and Kalyan]{lu2023dynamic}
Lu, P., Qiu, L., Chang, K.-W., Wu, Y.~N., Zhu, S.-C., Rajpurohit, T., Clark, P., and Kalyan, A.
\newblock Dynamic prompt learning via policy gradient for semi-structured mathematical reasoning.
\newblock In \emph{International Conference on Learning Representations (ICLR)}, 2023{\natexlab{c}}.

\bibitem[Lu et~al.(2023{\natexlab{d}})Lu, Qiu, Yu, Welleck, and Chang]{lu2022survey}
Lu, P., Qiu, L., Yu, W., Welleck, S., and Chang, K.-W.
\newblock A survey of deep learning for mathematical reasoning.
\newblock In \emph{The 61st Annual Meeting of the Association for Computational Linguistics (ACL)}, 2023{\natexlab{d}}.

\bibitem[McQuarrie(2008)]{mcquarrie2008quantum}
McQuarrie, D.~A.
\newblock \emph{Quantum chemistry}.
\newblock University Science Books, 2008.

\bibitem[Mishra et~al.(2022)Mishra, Finlayson, Lu, Tang, Welleck, Baral, Rajpurohit, Tafjord, Sabharwal, Clark, et~al.]{mishra2022lila}
Mishra, S., Finlayson, M., Lu, P., Tang, L., Welleck, S., Baral, C., Rajpurohit, T., Tafjord, O., Sabharwal, A., Clark, P., et~al.
\newblock Lila: A unified benchmark for mathematical reasoning.
\newblock In \emph{The 2022 Conference on Empirical Methods in Natural Language Processing (EMNLP)}, 2022.

\bibitem[OpenAI.(2022)]{chatgpt}
OpenAI.
\newblock Chatgpt: Optimizing language models for dialogue.
\newblock \url{https: //openai.com/blog/chatgpt/.}, 2022.

\bibitem[OpenAI.(2023)]{gpt4}
OpenAI.
\newblock Gpt-4 technical report.
\newblock \emph{arXiv preprint arXiv:2303.08774}, 2023.

\bibitem[Schick et~al.(2023)Schick, Dwivedi-Yu, Dess{\`\i}, Raileanu, Lomeli, Zettlemoyer, Cancedda, and Scialom]{schick2023toolformer}
Schick, T., Dwivedi-Yu, J., Dess{\`\i}, R., Raileanu, R., Lomeli, M., Zettlemoyer, L., Cancedda, N., and Scialom, T.
\newblock Toolformer: Language models can teach themselves to use tools.
\newblock \emph{arXiv preprint arXiv:2302.04761}, 2023.

\bibitem[Stewart et~al.(2012)Stewart, Watson, and Clegg]{stewart2012calculus}
Stewart, J., Watson, S., and Clegg, D.
\newblock Calculus: Early transcendentals, 8th.
\newblock \emph{Edition, Brooks/Cole, Cengae learning}, 2012.

\bibitem[Sun et~al.(2023)Sun, Han, Zhao, Ma, Shen, Chen, Chen, and Yu]{sun2023scieval}
Sun, L., Han, Y., Zhao, Z., Ma, D., Shen, Z., Chen, B., Chen, L., and Yu, K.
\newblock Scieval: A multi-level large language model evaluation benchmark for scientific research.
\newblock \emph{arXiv preprint arXiv:2308.13149}, 2023.

\bibitem[Suzgun et~al.(2022)Suzgun, Scales, Sch{\"a}rli, Gehrmann, Tay, Chung, Chowdhery, Le, Chi, Zhou, et~al.]{suzgun2022challenging}
Suzgun, M., Scales, N., Sch{\"a}rli, N., Gehrmann, S., Tay, Y., Chung, H.~W., Chowdhery, A., Le, Q.~V., Chi, E.~H., Zhou, D., et~al.
\newblock Challenging big-bench tasks and whether chain-of-thought can solve them.
\newblock \emph{arXiv preprint arXiv:2210.09261}, 2022.

\bibitem[Taylor et~al.(2022)Taylor, Kardas, Cucurull, Scialom, Hartshorn, Saravia, Poulton, Kerkez, and Stojnic]{taylor2022galactica}
Taylor, R., Kardas, M., Cucurull, G., Scialom, T., Hartshorn, A., Saravia, E., Poulton, A., Kerkez, V., and Stojnic, R.
\newblock Galactica: A large language model for science.
\newblock \emph{arXiv preprint arXiv:2211.09085}, 2022.

\bibitem[Thornton \& Marion(2021)Thornton and Marion]{thornton2021classical}
Thornton, S.~T. and Marion, J.~B.
\newblock \emph{Classical dynamics of particles and systems}.
\newblock Cengage Learning, 2021.

\bibitem[Touvron et~al.(2023{\natexlab{a}})Touvron, Lavril, Izacard, Martinet, Lachaux, Lacroix, Rozi{\`e}re, Goyal, Hambro, Azhar, et~al.]{llama}
Touvron, H., Lavril, T., Izacard, G., Martinet, X., Lachaux, M.-A., Lacroix, T., Rozi{\`e}re, B., Goyal, N., Hambro, E., Azhar, F., et~al.
\newblock {LLaMA}: Open and efficient foundation language models.
\newblock \emph{arXiv preprint arXiv:2302.13971}, 2023{\natexlab{a}}.

\bibitem[Touvron et~al.(2023{\natexlab{b}})Touvron, Martin, Stone, Albert, Almahairi, Babaei, Bashlykov, Batra, Bhargava, Bhosale, et~al.]{touvron2023llama}
Touvron, H., Martin, L., Stone, K., Albert, P., Almahairi, A., Babaei, Y., Bashlykov, N., Batra, S., Bhargava, P., Bhosale, S., et~al.
\newblock Llama 2: Open foundation and fine-tuned chat models.
\newblock \emph{arXiv preprint arXiv:2307.09288}, 2023{\natexlab{b}}.

\bibitem[Wang et~al.(2022)Wang, Wei, Schuurmans, Le, Chi, and Zhou]{wang2022self}
Wang, X., Wei, J., Schuurmans, D., Le, Q., Chi, E., and Zhou, D.
\newblock Self-consistency improves chain of thought reasoning in language models.
\newblock \emph{arXiv preprint arXiv:2203.11171}, 2022.

\bibitem[Wei et~al.(2022)Wei, Wang, Schuurmans, Bosma, Chi, Le, and Zhou]{wei2022chain}
Wei, J., Wang, X., Schuurmans, D., Bosma, M., Chi, E., Le, Q., and Zhou, D.
\newblock Chain of thought prompting elicits reasoning in large language models.
\newblock \emph{arXiv preprint arXiv:2201.11903}, 2022.

\bibitem[Welleck et~al.(2021)Welleck, Liu, Bras, Hajishirzi, Choi, and Cho]{welleck2021naturalproofs}
Welleck, S., Liu, J., Bras, R.~L., Hajishirzi, H., Choi, Y., and Cho, K.
\newblock Naturalproofs: Mathematical theorem proving in natural language.
\newblock \emph{arXiv preprint arXiv:2104.01112}, 2021.

\bibitem[Zhang et~al.(2024)Zhang, Hu, Zhoubian, Du, Yang, Wang, Yue, Dong, and Tang]{zhang2024sciglm}
Zhang, D., Hu, Z., Zhoubian, S., Du, Z., Yang, K., Wang, Z., Yue, Y., Dong, Y., and Tang, J.
\newblock Sciglm: Training scientific language models with self-reflective instruction annotation and tuning.
\newblock \emph{arXiv preprint arXiv:2401.07950}, 2024.

\bibitem[Zhang et~al.(2023{\natexlab{a}})Zhang, Han, Zhou, Hu, Yan, Lu, Li, Gao, and Qiao]{lamaadapter}
Zhang, R., Han, J., Zhou, A., Hu, X., Yan, S., Lu, P., Li, H., Gao, P., and Qiao, Y.
\newblock Llama-adapter: Efficient fine-tuning of language models with zero-init attention.
\newblock \emph{arXiv preprint arXiv:2303.16199}, 2023{\natexlab{a}}.

\bibitem[Zhang et~al.(2023{\natexlab{b}})Zhang, Zhang, Li, Zhao, Karypis, and Smola]{mmcot}
Zhang, Z., Zhang, A., Li, M., Zhao, H., Karypis, G., and Smola, A.
\newblock Multimodal chain-of-thought reasoning in language models.
\newblock \emph{arXiv preprint arXiv:2302.00923}, 2023{\natexlab{b}}.

\bibitem[Zhong et~al.(2023)Zhong, Cui, Guo, Liang, Lu, Wang, Saied, Chen, and Duan]{zhong2023agieval}
Zhong, W., Cui, R., Guo, Y., Liang, Y., Lu, S., Wang, Y., Saied, A., Chen, W., and Duan, N.
\newblock Agieval: A human-centric benchmark for evaluating foundation models.
\newblock \emph{arXiv preprint arXiv:2304.06364}, 2023.

\bibitem[Zhou et~al.(2022)Zhou, Sch{\"a}rli, Hou, Wei, Scales, Wang, Schuurmans, Bousquet, Le, and Chi]{zhou2022least}
Zhou, D., Sch{\"a}rli, N., Hou, L., Wei, J., Scales, N., Wang, X., Schuurmans, D., Bousquet, O., Le, Q., and Chi, E.
\newblock Least-to-most prompting enables complex reasoning in large language models.
\newblock \emph{arXiv preprint arXiv:2205.10625}, 2022.

\end{thebibliography}
